\documentclass[10pt,twocolumn,letterpaper]{article}

\usepackage[pagenumbers]{cvpr}

\graphicspath{{fig/}}

\usepackage{currfile} 
\usepackage{stfloats}

\usepackage{multirow}
\usepackage{longtable}
\usepackage{float}

\usepackage{tikz}
\usetikzlibrary{arrows.meta,positioning,fit,backgrounds,calc}

\definecolor{cvprblue}{rgb}{0.21,0.49,0.74}
\usepackage[pagebackref,breaklinks,colorlinks,allcolors=cvprblue]{hyperref}

\def\confName{CVPR}
\def\confYear{2027}

\title{FATE: Fault-Aware Transfer Error for Unsupervised Anomaly Detection}

\author{Tao Wang}

\begin{document}
\maketitle

\begin{abstract}
We study unsupervised industrial anomaly detection in the single-class setting, where a detector
learns only from normal images of one product category. Reconstruction-based detectors model
normal features from a frozen encoder, but their target layers are usually chosen by architectural
convention. We instead ask what information one block can pass to another. Treating depth as a
sequence of discrete transitions, we train eight directed, token-wise predictors from block 1 and
blocks 21--23 to blocks 7 and 14.

Adjacent-layer cosine similarity drops at four repeatable fault lines in DINOv2 ViT-L, separating
stable depth segments. The deepest blocks retain semantic context but lose local detail, so the
residual of a source-to-sensitive prediction measures information that cannot be transported. The
fault structure constrains the source and target roles, and the same rule transfers to ViT-B and
RADIO-B.

A bounded foreground prior, calibrated only on normal training images, suppresses background
residuals while retaining a positive floor for evidence outside the estimated foreground. With the
same eight channels and a common training-only calibration procedure in every single-category
model, category-averaged image AUROC reaches $99.75$ on MVTec-AD, $99.42$ on VisA and $94.27$ on
Real-IAD; object-level AUROC on Real-IAD is $96.26$. For the $448^2$ runs, category-averaged
P-AUROC/P-AUPRO on a common $256^2$ evaluation grid is $98.46/95.97$ on MVTec-AD, $99.21/96.12$ on
VisA and $99.26/96.59$ on Real-IAD.
\end{abstract}

\begin{figure*}[!b]
\centering
\includegraphics[width=0.94\textwidth]{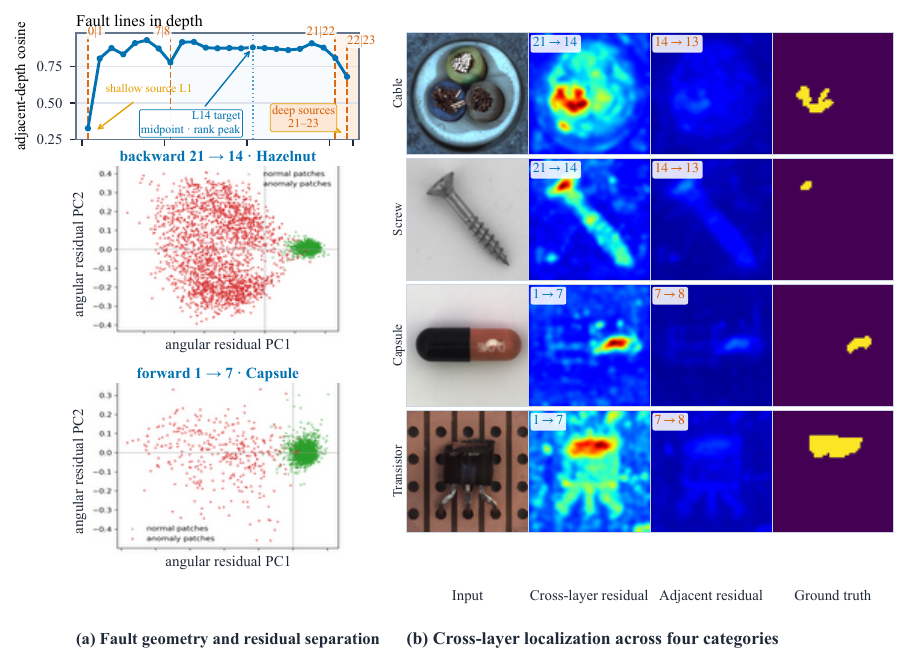}
\caption{\textbf{Measured evidence for fault-guided cross-layer prediction.}
(\textbf{a}) Adjacent-depth cosine similarity averaged over five categories. Four faults separate
stable depth segments. L14 is marked as the midpoint of the stable segment and the
participation-ratio peak. The two angular-residual manifolds show normal patches concentrating
near the origin while defect patches spread outward for backward $21{\to}14$ and forward $1{\to}7$.
(\textbf{b}) One anomaly from each of four MVTec-AD
categories. The first
two rows use $21{\to}14$ and the last two use $1{\to}7$, illustrating complementary backward and
forward evidence; the adjacent column compares these with $14{\to}13$ and $7{\to}8$ under the same
training run.}
\label{fig:teaser}
\end{figure*}

\section{Introduction}
\label{sec:intro}

Industrial inspection must detect defects that cannot be enumerated in advance. We consider the
single-class setting, where training uses only normal images from one product category.
MVTec-AD~\citep{bergmann2019mvtec}, VisA~\citep{zou2022visa} and
Real-IAD~\citep{wang2024realiad} follow this protocol. The leading methods reconstruct features
from a frozen encoder~\citep{deng2022rd4ad,guo2025dinomaly,guo2025dinomaly2}. A bottleneck
compresses the features, a decoder reconstructs selected teacher features, and anomalies appear as
large residuals. Fig.~\ref{fig:teaser} motivates a different design: the deepest features remain insensitive to local defects,
while an asymmetric cross-layer predictor tests whether deep and shallow sources can transport
the missing evidence to sensitive blocks; the transfer failure becomes the anomaly signal.

Reconstruction methods share two assumptions. First, the decoder can see the features it must
reproduce, which permits a copying shortcut. Second, the target representations are selected by
architectural convention. RD4AD~\citep{deng2022rd4ad} reconstructs multi-scale encoder
representations through a one-class bottleneck, whereas the Dinomaly
line~\citep{guo2025dinomaly,guo2025dinomaly2} reconstructs a fixed group of ViT blocks.
Neither choice is tied to a measured property of the representation, and sensitive and insensitive
layers are treated alike.

We test the backbone directly. Adjacent-layer cosine similarity is not smooth in depth. In DINOv2
ViT-L it drops sharply at $0|1$, $7|8$, $21|22$ and $22|23$, partitioning the 24 blocks into
$[1..7]$, $[8..21]$, $[22]$ and $[23]$, with block $0$ isolated before the first fault. Effective
feature rank and the response to a local perturbation change at the same boundaries. The deepest
blocks therefore carry semantic context but little local defect detail. These signatures are
consistent with rank-collapse analyses of deep attention and transformer
stacks~\citep{dong2021pureattention,noci2022signalprop}. The layer-wise phase transitions in ViTs
are consistent with terminal neural-collapse geometry~\citep{krishnakumar2025cliffplateau}, building
on the general neural-collapse phenomenon~\citep{papyan2020neuralcollapse}.

We treat depth as a sequence of transitions~\citep{aswani2026dmdvit,jacobs2025blockrecurrent,
shopkhoev2025replaceme}. A directed channel $(s,t)$ learns
$g_{s\to t}$ from block $s$ to block $t$, while the backbone remains frozen. The residual
$1-\cos(g_{s\to t}(f_s),f_t)$ is the part of the target state that the source cannot transport.
The source and target therefore become explicit design choices. A deep block is useful as a source
because it lacks local detail; a sensitive block is useful as a target because that missing detail
becomes visible there.

The fault structure determines the roles. Block $1$ is the shallow source, block $7$ is the near
target before the $7|8$ fault, block $14$ is the midpoint of $[8..21]$, and blocks $21$--$23$ are
deep sources. Fig.~\ref{fig:target_selection} checks this assignment against ground-truth and
synthetic anomalies. The curves remain sensitive at L7 and L14 and decay in the final segment. L7
remains the near target, while L14 supplies the mid-depth target. Each channel is scored
independently, then the eight scores are averaged. A soft foreground prior can attenuate background
residuals but never sets them to zero.

Directed transfer between selected patch embeddings was used by
Costanzino et al.~\citep{costanzino2025tinydefects}, with layer pairs chosen by ablation. We keep
the directed transfer and derive the pairs and their direction from the backbone's fault structure.
The same rule selects layers for ViT-L, ViT-B and RADIO-B.

Our contributions are:
\begin{itemize}
  \item a label-free layer-selection rule that detects adjacent-layer cosine faults, corroborated
  by effective-rank and local-sensitivity diagnostics;
  \item a directed cross-layer predictor that turns asymmetric source-to-sensitive transfer
  failure into an anomaly score, with deep-to-sensitive channels as the principal signal and
  shallow-to-sensitive channels as complementary evidence;
  \item a shared scoring and foreground-prior procedure that transfers across three backbones and
  three industrial benchmarks without per-category channel selection.
\end{itemize}

With the same predictor architecture and training-only calibration, image AUROC is $99.75$ on
MVTec-AD, $99.42$ on VisA and $94.27$ on Real-IAD. Real-IAD object-level AUROC is $96.26$;
at $256\times256$ pixel resolution, P-AUROC/P-AUPRO is $99.26/96.59$.

\section{Related Work}
\label{sec:related}

\textbf{Reconstruction and distillation.} RD4AD~\citep{deng2022rd4ad} reconstructs multi-scale
frozen encoder representations through a one-class bottleneck. The Dinomaly
line~\citep{guo2025dinomaly,guo2025dinomaly2} reconstructs a selected group of ViT blocks.
DeSTSeg~\citep{zhang2023destseg} adds segmentation-guided denoising, while unified models alter
their architecture, bottleneck, objective or aggregation~\citep{you2022uniad,guo2023recontrast,
zhang2025vitad,he2024mambaad}. These methods condition the decoder on the features it reconstructs and choose the
target representations by convention. Our predictor maps one layer to a different layer, and the
measured fault structure defines the source and target roles.

\textbf{Directed prediction.} Costanzino et al.~\citep{costanzino2025tinydefects} train forward and
backward MLPs between a benchmark-selected layer pair and multiply their residuals as a
bidirectional fail-safe, assuming that the selected inter-layer transition is invertible. Our
measurements instead expose a segmented regime. The $0|1$ fault isolates the initial block and
identifies block $1$ as the first stable shallow source, while participation collapses across the
final segment after $21|22$, with further compression at $22|23$. The deepest blocks therefore no
longer carry the fine local evidence needed to predict sensitive targets faithfully. We use both
shallow and deep blocks as sources and sensitive blocks as targets. The fault structure determines
these roles rather than benchmark ablation: L7 is the last sensitive block before the $7|8$ fault,
L14 is the midpoint of $[8..21]$ with a participation-ratio tie-break, and block $0$ is not used as
a production endpoint. Forward and backward predictors remain distinct directed channels rather
than being collapsed into a single symmetric bidirectional fail-safe design.

\textbf{Embedding and synthesis methods.} PatchCore~\citep{roth2022patchcore} scores a coreset of
normal patch features, and PaDiM~\citep{defard2021padim} models each spatial position with a
Gaussian. CutPaste~\citep{li2021cutpaste}, DRAEM~\citep{zavrtanik2021draem},
SimpleNet~\citep{liu2023simplenet}, GLASS~\citep{chen2024glass} and
EfficientAD~\citep{batzner2024efficientad} use synthetic anomalies, discriminative training or a
lightweight student. These methods do not model directed relations along depth. A more detailed
discussion of related work is provided in Appendix~\ref{sec:suppl_related}.

\begin{figure*}[t]
\centering
\includegraphics[width=\linewidth]{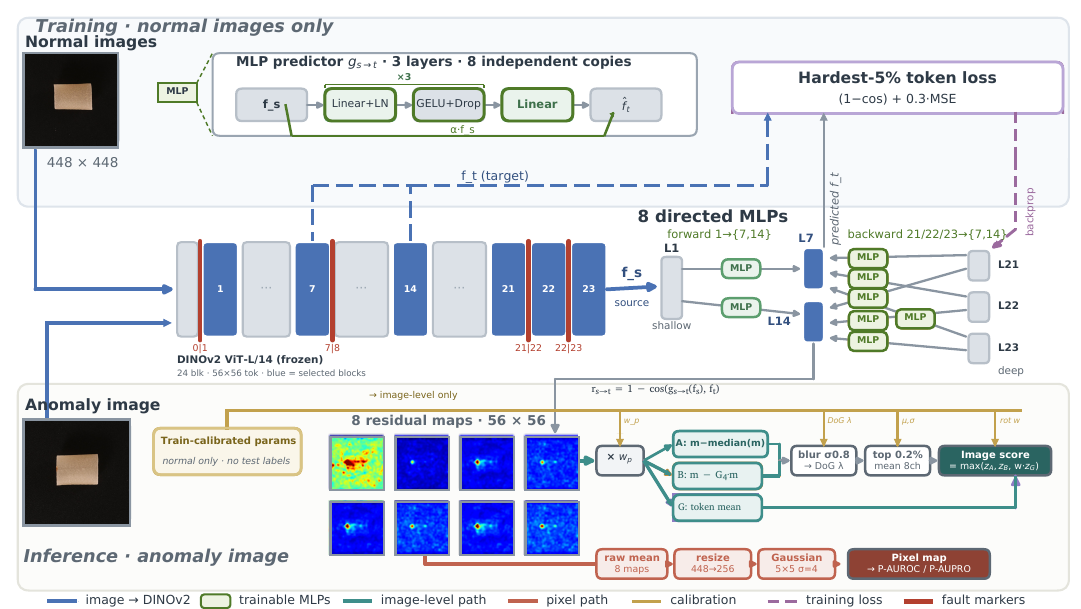}
\caption{Pipeline overview. Training and inference share a frozen DINOv2 ViT-L/14 and eight
directed predictors. Four cosine faults select six blocks; each MLP maps one source block to one
target, and the loss averages the hardest 5\% of tokens. The residual
$r_{s\to t}=1-\cos(g_{s\to t}(f_s),f_t)$ measures what the source cannot carry, and the predictor
retains a trainable source shortcut $\alpha_{s\to t}f_s$. At test time, a training-only soft prior
weights the residuals. The image score uses
$\max(z_A,z_B,wz_G)$ after median/spatial baselines, smoothing, high-pass filtering and
top-0.2\% pooling. Pixel localization bypasses these image-level branches and is evaluated at
$256\times256$ resolution as the direct mean of the eight residual maps.}
\label{fig:framework}
\end{figure*}

\section{Method}
\label{sec:method}

\subsection{Overview}
\label{sec:overview}

Given the normal images of one category, the pipeline contains a frozen feature extractor, eight
directed predictors and a label-free scoring stage (Fig.~\ref{fig:framework}). Every reported model
is trained separately for one category; no multi-class objective or cross-category predictor
parameters are used. We use DINOv2
ViT-L/14 with registers~\citep{oquab2024dinov2,darcet2024registers}. Its parameters remain frozen.
At $448^2$, the $14\times14$ patch embedding uses stride $8$ and padding $3$, producing an
overlapping $56\times56$ token grid ($3136$ tokens); the positional embedding is interpolated to
the same grid. We keep blocks $\mathcal{L}=\{1,7,14,21,22,23\}$ from the 24-block backbone
(zero-based indices).

Each channel is an ordered pair $(s,t)$, $s\neq t$, implemented by a separate three-layer MLP with
hidden width $512$. A learnable source shortcut preserves information that lies close to $f_s$:
\begin{equation}
\hat f_t \;=\; g_{s\to t}(f_s) \;=\; \mathrm{MLP}(f_s) \;+\; \alpha_{s\to t} f_s ,
\label{eq:predictor}
\end{equation}
where $\alpha_{s\to t}$ is initialised to $0.1$ and trained jointly with the MLP. Source and target
features have the same width in all evaluated backbones.
The forward and backward pairs are listed in Table~\ref{tab:pairs}.

\begin{table}[t]
\centering
\small
\begin{tabular}{lll}
\toprule
branch & source block $s$ & target blocks $t$ \\
\midrule
forward  & $1$            & $14,\;7$ \\
backward & $21,\;22,\;23$ & $14,\;7$ \\
\bottomrule
\end{tabular}
\caption{The eight directed channels. Both branches are trained jointly; the pairing rule
is derived in Sec.~\ref{sec:faults} and the role of each direction is analysed in
Sec.~\ref{sec:crosslayer}.}
\label{tab:pairs}
\end{table}

\textbf{Objective and optimisation.} For prediction $\hat f_t = g_{s\to t}(f_s)$, the token loss is
\begin{equation}
\ell_t^i \;=\; \bigl(1-\cos(\hat f_t^i, f_t^i)\bigr)
\;+\; \frac{0.3}{D}\bigl\|\hat f_t^i - f_t^i\bigr\|_2^2 ,
\label{eq:tokenloss}
\end{equation}
and the channel averages $\ell$ over the hardest $5\%$ of tokens. This prevents well-modelled
normal tokens from dominating a small defect. The loss averages within each direction and then
weights the forward and backward branch means equally; an individual forward channel therefore has
three times the weight of a backward channel. All coefficients are fixed across
categories, datasets and backbones. We optimise only the predictors with AdamW (weight decay
$10^{-4}$, learning rate $3\times10^{-3}$, cosine schedule, feature batch size
$8$)~\citep{loshchilov2019adamw}. The epoch
budget is $\mathrm{clip}(\mathrm{round}(N_{\mathrm{train}}/8),10,40)$; training stops after two
epochs without $0.1\%$ relative improvement, with a minimum of five epochs.

\subsection{Fault Lines in the Backbone}
\label{sec:faults}

We select layers without labels, masks or test images. For training images, let $\bar c(\ell)$ be
the mean cosine similarity between block $\ell$ and block $\ell{+}1$ over positions and images.
Across five categories and three datasets, it drops at four repeatable fault lines:
\begin{equation}
\bar c(\ell)\;\approx\;
\begin{cases}
0.28\text{--}0.39, & \ell=0 \quad (0|1),\\
0.77\text{--}0.79, & \ell=7 \quad (7|8),\\
0.80\text{--}0.81, & \ell=21 \quad (21|22),\\
0.65\text{--}0.69, & \ell=22 \quad (22|23),
\end{cases}
\end{equation}
while the remaining boundaries lie between $0.85$ and $0.93$. We detect faults with relative tests
and shared constants $\delta$ and $k$. A boundary $\ell=1,\dots,21$ is a fault if
\begin{equation}
\bar c(\ell) \;<\; \frac{\bar c(\ell-1)+\bar c(\ell+1)}{2} - \delta
\end{equation}
or if the drop into it is large relative to the whole sequence, for $\ell=1,\dots,22$,
\begin{equation}
\bar c(\ell-1)-\bar c(\ell) \;>\; k\,\mathrm{std}\bigl(\bar c\bigr).
\end{equation}
The left endpoint $0|1$ uses the same drop test in the opposite direction,
\begin{equation}
\bar c(1)-\bar c(0) \;>\; k\,\mathrm{std}\bigl(\bar c\bigr).
\end{equation}
For ViT-L, $\delta=0.08$ and $k=0.5$; ViT-B and RADIO use the same rule. The faults partition the
24 blocks into four stable segments
\begin{equation}
[1..7],\qquad [8..21],\qquad [22],\qquad [23],
\end{equation}
with block $0$ isolated before the first fault. Two label-free diagnostics follow the same
boundaries. The participation ratio peaks in the middle segment (34.5 at block 14 for
\texttt{cable}) and falls to 10.7 at block 23. A synthetic local perturbation also produces a
weaker and broader response in the deepest blocks. These signatures agree with rank-collapse
analyses of deep attention and transformer stacks~\citep{dong2021pureattention,noci2022signalprop}.
The layer-wise phase transitions in ViTs further agree with terminal neural-collapse
geometry~\citep{krishnakumar2025cliffplateau,papyan2020neuralcollapse}.

The layer sequence can also be viewed through a discrete-time analogy: block-recurrent,
linearized, and dynamic-mode analyses model relations across ViT
depth~\citep{aswani2026dmdvit,shopkhoev2025replaceme,jacobs2025blockrecurrent}, while Koopman,
neural-ODE, and stable-architecture formulations provide the broader dynamical-systems
language~\citep{colbrook2024rigorous,patyn2024koopman,chen2018neuralode,haber2017stable}. We use
this only as an analysis analogy; each deployed transition remains a separately trained pointwise
MLP. The rule transfers directly to 12-block backbones: ViT-B and RADIO-B both show a shallow notch
at $3|4$ and a late collapse, giving near target $3$, mid target $7$, and deep sources $10,11$
(Table~\ref{tab:backbone_faults}). Only the layer indices change; the predictors, scoring and prior
rules remain the same.

\begin{table}[t]
\centering
\scriptsize
\setlength{\tabcolsep}{1.5pt}
\begin{tabular}{@{}p{0.17\columnwidth}p{0.29\columnwidth}ccp{0.17\columnwidth}@{}}
\toprule
backbone & detected faults & near target & mid target & deep source band \\
\midrule
\mbox{ViT-L}  & $0|1$, $7|8$, $21|22$, $22|23$ & $7$ & $14$ & $21,22,23$ \\
\mbox{ViT-B}  & $3|4$, $9|10$ & $3$ & $7$ & $10,11$ \\
\mbox{RADIO-B} & $3|4$, $9|10$ & $3$ & $7$ & $10,11$ \\
\bottomrule
\end{tabular}
\captionsetup{font=footnotesize}
\caption{Fault boundaries produced by the relative rule on the recorded cosine curves,
together with the layer set they imply. The ViT-B and RADIO-B rows are the configurations used
in the completed backbone-generalization experiments.}
\label{tab:backbone_faults}
\end{table}

\subsection{Directed Cross-Layer Prediction}
\label{sec:crosslayer}

Reconstruction residuals mix two effects: information lost in the bottleneck and the ability of the
teacher representation to reproduce itself. A directed channel avoids the second effect. With depth
treated as discrete time, the hidden states form a trajectory, and $g_{s\to t}$ learns the
transition from block $s$ to block $t$ pointwise over tokens. At inference, the residual
$1-\cos(\hat f_t,f_t)$ measures the part of the target state that the source cannot transport. We
use this as an analysis abstraction, not as a formal claim that a transformer block is an autonomous
dynamical system.

\subsection{Layer-Role Diagnostics}
\label{sec:layer_roles}

\begin{figure}[t]
\centering
\includegraphics[width=\columnwidth]{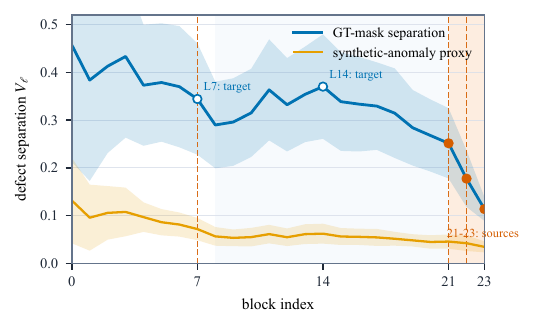}
\caption{Source/target placement from layer-wise defect separation. Blue uses ground-truth masks;
orange uses synthetic anomalies. Shading shows $\pm1$ standard deviation over five categories.
L7 and L14 remain sensitive and serve as targets; separation falls over blocks 21--23, which serve
as sources. L7 is kept as the near target because it is the last sensitive block before $7|8$,
not because it is the global maximum.}
\label{fig:target_selection}
\end{figure}

Two diagnostics, neither used at inference, check the inferred roles. For layer $\ell$,
\begin{align}
d_\ell(x,p)
&= 1-\cos\!\left(f_\ell(x,p),\,\bar f_\ell^{\mathrm{train}}(p)\right),\\
V_\ell
&= \operatorname{mean}_{p\in\mathcal{D}}\!\left[d_\ell(x,p)\right] \notag\\
&\quad -\operatorname{median}_{p\in\mathcal{N}}\!\left[d_\ell(x,p)\right],
\end{align}
where $\bar f_\ell^{\mathrm{train}}(p)$ is the mean training feature at position $p$. The sets
$\mathcal{D}$ and $\mathcal{N}$ contain defect and normal positions. We compute $V_\ell$ once with
test masks for diagnosis and once with synthetic occlusions as a label-free proxy. Both curves fall
over blocks 21--23. Fig.~\ref{fig:target_selection} therefore keeps L7 and L14 as targets and
21--23 as sources. L7 is slightly weaker than L14 but sits immediately before the $7|8$ fault, so
it remains the near target; L14 is the mid-depth target. Participation ratio gives the same
picture: the middle layers retain many feature directions, while the late layers become compact
and insensitive.

For a segment $[a..b]$, the mid target is the nearest integer to $(a+b)/2$, with ties broken by
participation ratio; this gives L14 for $[8..21]$ in ViT-L and L7 for $[4..9]$ in the 12-block
backbones. The faults constrain the role regions rather than uniquely determine every endpoint. We
fix block 1 as the shared shallow source, a design convention rather than a claim of uniqueness.
Targets are fixed without test labels, and the ablation below tests nearby targets and alternative
source counts.

\subsection{Scoring: Dual Per-Image Baseline and Train-Calibrated High-Pass}
\label{sec:scoring}

Let $m$ be a channel's $56\times56$ residual map after multiplication by the soft-prior weight
$w_p$ from Eq.~\eqref{eq:softweight}. We score the map with two complementary per-image baselines.
Branch A removes a global offset using the median over all tokens:
\begin{equation}
m_A \;=\; m - \mathrm{median}(m) .
\end{equation}

Branch B removes a smooth spatial field estimated by a $\sigma=4$ Gaussian:
\begin{equation}
m_B \;=\; m - G_{4}\!\ast m .
\end{equation}
This keeps narrow peaks from scratches and small holes while suppressing illumination drift. Both
maps are then smoothed with a $3\times3$ Gaussian ($\sigma=0.8$) and high-pass filtered:
\begin{equation}
m_* \;\leftarrow\; m_* - \lambda\,G_{\sigma}\!\ast m_* , \qquad \sigma = 2.0 ,
\end{equation}
The mean of the largest $0.2\%$ patches ($k\geq5$) gives one channel score. The eight channel scores
are averaged separately for A and B. Smoothing removes pixel noise; the high-pass suppresses
low-frequency drift. Its strength $\lambda$ is computed once per category from the first 25 normal
training images in deterministic loader order:
\begin{equation}
\rho \;=\; \frac{\sum |G_{\sigma}\!\ast m|}{\sum |m|} ,
\end{equation}
followed by
\begin{equation}
\lambda \;=\; 0.75 \cdot \mathrm{clip}\!\left(\frac{\rho-0.80}{0.88-0.80},\,0,\,1\right) .
\end{equation}

Table~\ref{tab:constants} lists the fixed inference values and calibration rules.

\begin{table}[t]
\centering
\scriptsize
\setlength{\tabcolsep}{2pt}
\begin{tabular}{@{}p{0.28\columnwidth}p{0.12\columnwidth}p{0.5\columnwidth}@{}}
\toprule
quantity & symbol & value \\
\midrule
high-pass strength   & $\lambda$ & $0.75\cdot\mathrm{clip}((\rho-0.80)/0.08,0,1)$ \\
high-pass scale      & $\sigma$  & $2.0$ \\
smoothing            & --        & $3\times3$ Gaussian, $\sigma=0.8$ \\
pooling ratio        & --        & top $0.2\%$ ($k\geq5$) \\
B-branch field       & $\sigma_B$ & $4$ \\
global reliability    & $w$       & $\exp(-\mathrm{rot}/0.30)$; $w=0$ for no mask or multi-object \\
multi-object test     & --        & connected components $n_{\mathrm{comp}}>2$ \\
prior floor          & $f$       & $0.7$ if multi-object; else $0.1$ if $BC\ge0.40$, else $0.7$ \\
mask-off             & --        & prior not applied \\
\bottomrule
\end{tabular}
\captionsetup{font=footnotesize}
\caption{Fixed inference values and training-only calibration rules. Thresholds are shared;
$\lambda$, normalisation and the mask prior use only normal training images.}
\label{tab:constants}
\end{table}

When normal residuals are already high frequency, $\rho<0.80$ and the filter switches off. A
low-frequency normal residual increases $\lambda$ up to $0.75$. The per-category values appear in
the supplement.

\subsection{Label-Free Soft Fusion}
\label{sec:fusion}

Lighting and contrast vary from image to image, so the correction has to be estimated per image.
The A and B branches provide the two baselines above. A third global branch uses the soft-mask-weighted
token mean of the residual and responds to whole-object shifts. We convert each branch to a
$z$-score using the first 60 normal training images in deterministic loader order, then fuse:
\begin{equation}
\begin{aligned}
s_{\mathrm{img}} &= \max\bigl(z_A,\; z_B,\; w \cdot z_{\mathrm{global}}\bigr),\\
w &= \exp\bigl(-r / \tau\bigr),
\end{aligned}
\label{eq:softfusion}
\end{equation}
where $\tau=0.30$ and $w$ is separate from the token weight $w_p$. To estimate pose rigidity $r$,
we compute the foreground principal-axis angle of each training SAM-2 mask, double it to remove the
$180^\circ$ sign ambiguity, and take the circular standard deviation. Rotating or texture
categories therefore turn the global branch off. For multi-object categories
($\mathrm{ncomp}>2$), the global mean mixes several objects and often anti-correlates with a
single-object defect, so we also set $w=0$ and retain floor $0.7$.

\subsection{Pixel-Level Localization Score}
\label{sec:pixelfusion}

The pixel metric deliberately bypasses the image-level branches: it does not apply the foreground
prior, the A/B median and spatial baselines, global fusion, DoG filtering or branch $z$-scores.
Instead, the localization map is the equal mean of the eight raw directed $56\times56$ residual
maps; it is rasterized to $448^2$, resized to $256^2$, then smoothed by a $5\times5$ Gaussian with
$\sigma=4$ and zero padding. Ground truth is resized directly to the same $256^2$ grid, and
P-AUROC and P-AUPRO are computed over all test pixels. This common $256\times256$ evaluation grid
preserves the spatial defect response without adding a second, separately tuned localization
pipeline.

\subsection{Soft Foreground Prior}
\label{sec:maskhead}

Unlike reconstruction, the pointwise residual has no bottleneck that compresses background
variation. Foreground filtering is therefore part of the detector rather than an optional
post-processing add-on. We train a foreground probe from SAM~2~\citep{ravi2025sam2} masks on normal
training images. For the six blocks in
$\mathcal{L}$, we use up to 50 cached masks in deterministic loader order, sample 800 patches per
image with a fixed seed, fit an L2 logistic regression per block ($C=1$), and keep the three blocks
with the highest separation. A joint L2 probe on their concatenated features gives a foreground
probability $p$ per patch. The cap keeps calibration cost fixed when Real-IAD categories contain
more than 1,000 normal images.

Before use, masks whose mean four-corner foreground coverage exceeds $0.5$ are polarity-flipped;
this guard is active for one VisA category. A separate fallback disables the probe when its
training-side pseudo-mask AUC is below $0.60$, and fewer than four cached masks also disable it.
The AUC fallback did not trigger on the reported benchmarks. All checks use training images and
pseudo-masks only.

The probe is trained on pseudo-masks, so $p$ acts as a prior rather than a decision. The patch
weight is
\begin{equation}
w_p \;=\; f + (1-f)\,p ,
\label{eq:softweight}
\end{equation}
so background patches retain at least a fraction $f$ of their residual. The prior can attenuate
evidence but cannot erase a misclassified defect.

The floor depends on how often the model's top residual patches fall outside the predicted
foreground. Let $T$ contain the top-eight residual patches and let $\mathcal{F}$ be the predicted
foreground set:
\begin{align}
BC &= \frac{|\{t \in T : t \notin \mathcal{F}\}|}{|T|},\\
f &=
\begin{cases}
0.7, & \mathrm{ncomp} > 2,\\
0.1, & BC \ge 0.40,\\
0.7, & \text{otherwise},
\end{cases}
\label{eq:bcfloor}
\end{align}
averaged over the first 25 normal training images and both branches. A high $BC$ can reflect either
background distractors or an incomplete mask, which training data alone cannot distinguish. We use
the contrastive floor $f=0.1$ only for accepted single-object masks; multi-object and unreliable
masks retain $f=0.7$. The fixed $0.40$ threshold is an operational calibration, not an optimality
result; the supplement lists its category decisions and calibration scope.

\subsection{Reliability Gate}
\label{sec:gate}

Pseudo-masks can miss a thin object or include textured background. From the first 25 training masks we
compute foreground fraction, connected components, aspect ratio and foreground/background variance.
The prior remains on for multi-object categories. Otherwise it is disabled when either rule holds:
\begin{align}
\text{rule A}:&\quad
\mathrm{ncomp}\le1.5 \;\wedge\; \mathrm{ar}\ge2.5 \;\wedge\; \mathrm{fg}<0.25,\notag\\
\text{rule B}:&\quad
\frac{\mathrm{var}_{\mathrm{fg}}}{\mathrm{var}_{\mathrm{bg}}} \le 1.0 .
\end{align}
Rule A catches a thin, small object with an implausible mask. Rule B catches masks with no usable
foreground/background contrast. Both use training images and pseudo-masks only; the disabled
categories are listed in the supplement.

\section{Experiments}
\label{sec:exp}

\subsection{Experimental Settings}
\label{sec:setup}

\textbf{Datasets and experimental protocol.} We evaluate on MVTec-AD~\citep{bergmann2019mvtec} (15 categories),
VisA~\citep{zou2022visa} (12), and Real-IAD~\citep{wang2024realiad} (30), training one model per
category, averaging category scores, and also grouping Real-IAD by object across its five
views (O-ROC).

\textbf{Implementation.} All datasets use a frozen DINOv2 ViT-L/14 at $448^2$ with images enhanced
at their native aspect ratio and resized directly to $448^2$ without letterbox padding, while input
synthesis and target scratch/noise synthesis are disabled, training and optimisation follow
Sec.~\ref{sec:method}, the backbone, predictor architecture, channel pairs, loss weights, floor
rule, and gate thresholds are shared, and category-specific calibration statistics for $\lambda$,
branch normalisation, and the mask prior are computed only from normal training images.

\textbf{Metrics.} We use image-level AUROC as the primary score (I-ROC on Real-IAD), compute the
pixel map for the refreshed $448^2$ results as the equal mean of the eight raw residual maps before
foreground weighting or image-level branch calibration, rasterized to $448^2$, resized to
$256^2$, and smoothed by a $5\times5$ Gaussian with $\sigma=4$ with ground truth evaluated on the
same $256^2$ grid, and report
P-AUROC and P-AUPRO with the AUPRO false-positive limit at $0.3$ from deterministic single runs,
while category-level dispersion is reported in the supplement and training-seed variance is not
estimated.

Additional portability tests on MPDD~\citep{jezek2021mpdd}, BTAD~\citep{mishra2021btad},
DAGM~\citep{wieler2007dagm}, and Kolektor Surface-Defect Dataset~2
(KSDD2)~\citep{bozic2021ksdd2} are reported in
Appendix~\ref{sec:suppl_extra_datasets}.

\subsection{Comparison Results}
\label{sec:comparison}

We report the official Real-IAD low-resolution benchmark, a Dinomaly-series comparison, and a
backbone transfer test separately, since they answer different questions.

\begin{table}[t]
\centering
\scriptsize
\setlength{\tabcolsep}{2pt}
\begin{tabular}{@{}p{0.48\columnwidth}ccc@{}}
\toprule
method & S-AUROC & I-AUROC & P-AUPRO \\
\midrule
PaDiM~\citep{defard2021padim}        & 91.2 & 80.3 & 81.8 \\
CFlow~\citep{gudovskiy2022cflow}     & 89.8 & 82.5 & 90.5 \\
PatchCore~\citep{roth2022patchcore}  & 93.7 & 89.4 & 91.5 \\
SoftPatch~\citep{jiang2022softpatch} & 92.8 & 89.4 & 90.8 \\
SimpleNet~\citep{liu2023simplenet}   & 94.9 & 88.5 & 84.6 \\
DeSTSeg~\citep{zhang2023destseg}     & 94.0 & 86.9 & 81.5 \\
RD~\citep{deng2022rd4ad}             & 83.7 & 87.1 & 93.8 \\
UniAD~\citep{you2022uniad}           & 88.1 & 82.9 & 86.1 \\
\midrule
\textbf{Ours, ViT-L, $252^2$}        & 95.10 & 92.21 & 93.86 \\
\bottomrule
\end{tabular}
\captionsetup{font=footnotesize}
\caption{Real-IAD reference benchmark. S, I and PRO are sample AUROC, image AUROC and P-AUPRO.
The first eight rows use the official 256-side protocol; our patch-aligned $252^2$ run is an
approximate resolution match. This table is used for benchmark-level calibration, not as a
strict protocol-identical claim.}
\label{tab:realiad-256}
\end{table}

Table~\ref{tab:realiad-256} is an approximate $252^2$/$256$ protocol comparison, not a
state-of-the-art ranking; newer single-class methods use different Real-IAD protocols and are
omitted. Table~\ref{tab:dinomaly-comparison} controls backbone, setting and resolution.

\begin{table*}[!t]
\centering
\small
\setlength{\tabcolsep}{3pt}
\renewcommand{\arraystretch}{0.95}
\begin{tabular}{ll cccc cccc}
\toprule
& & \multicolumn{4}{c}{$392^2$} & \multicolumn{4}{c}{$448^2$} \\
\cmidrule(lr){3-6} \cmidrule(lr){7-10}
dataset & method & I & O & P & PRO & I & O & P & PRO \\
\midrule
\multirow{2}{*}{MVTec-AD} & ours & 99.70 & -- & 97.30 & 95.94 & 99.75 & -- & 98.46 & 95.97 \\
                          & Dinomaly & 99.84 & -- & 98.54 & 95.83 & 99.85 & -- & 98.75 & 96.42 \\
\midrule
\multirow{2}{*}{VisA} & ours & 99.29 & -- & 98.95 & 94.83 & 99.42 & -- & 99.21 & 96.12 \\
                      & Dinomaly & 99.05 & -- & 99.21 & 95.90 & 98.90 & -- & 99.39 & 96.64 \\
\midrule
\multirow{2}{*}{Real-IAD} & ours & 94.01 & 96.13 & 98.70 & 97.33 & 94.27 & 96.26 & 99.26 & 96.59 \\
                         & Dinomaly & 92.67 & 96.61 & 99.24 & 96.26 & 91.98 & 96.63 & 99.43 & 96.79 \\
\bottomrule
\end{tabular}
\captionsetup{font=small}
\caption{Comparison with the Dinomaly series. All entries are single-class, per-category runs; the
$392^2$ rows are our preceding runs and the $448^2$ rows use the refreshed geometry. We do not use
the official multi-class benchmark here. I and O are image- and object-level AUROC; P and PRO are
P-AUROC and P-AUPRO. The Dinomaly series uses its native center crop, whereas the refreshed ours
run resizes the full image directly to $448^2$; its pixel maps and ground truth are compared on a
common $256\times256$ grid.}
\label{tab:dinomaly-comparison}
\end{table*}

Under the refreshed $448^2$ protocol, our method leads on VisA image AUROC by $0.52$~pp and on
Real-IAD image AUROC by $2.29$~pp. The Dinomaly series leads on MVTec-AD image AUROC by $0.10$~pp,
Real-IAD object AUROC by $0.37$~pp, and both pixel metrics on all three datasets. The result is
therefore an image-level advantage on two benchmarks, not a universal pixel-localization win.
The $392^2$ row is retained from the preceding matched run; only the $448^2$ benchmark set was
retrained for the refreshed geometry.

\subsection{Ablation Study}
\label{sec:ablation}
We ablate on four categories (cable, screw, pcb1, macaroni2) with the production
soft-gated scoring. The matched reference configuration has mean image AUROC $0.9902$; every row of
Table~\ref{tab:ablation} changes one component while everything else stays fixed. These mechanism
controls use the preceding locked checkpoints; the main $448^2$ table is the independently
refreshed run, so the ablation is interpreted within its matched family rather than as an absolute
delta to the refreshed score.

\begin{table}[t]
\centering
\footnotesize
\setlength{\tabcolsep}{2.5pt}
\renewcommand{\arraystretch}{0.92}
\begin{tabular}{@{}p{0.58\columnwidth}cc@{}}
\toprule
configuration & mean AUROC & $\Delta$ (pp) \\
\midrule
reference (8 predictors, B3)           & 0.9902 & -- \\
\emph{branch group} & & \\
forward channels only                  & 0.9663 & $-2.39$ \\
all backward channels (BWD6)           & 0.9874 & $-0.28$ \\
\emph{targets} & & \\ 
drop target $L_7$ (ONLY14)             & 0.9866 & $-0.36$ \\
same-source $14{\to}7$ test (B7)       & 0.9626 & $-2.76$ \\
targets in blind/far depth (FAR4)      & 0.9386 & $-5.17$ \\
target block $L_0$ (TGT0)              & 0.9795 & $-1.07$ \\
\emph{sources} & & \\ 
one deep source $\{23\}$ (B1)          & 0.9873 & $-0.29$ \\
two deep sources $\{22,23\}$ (B2)      & 0.9892 & $-0.10$ \\
four deep sources $\{20,21,22,23\}$ (B4) & 0.9902 & $0.00$ \\
deep sources $\{8,9,10\}$              & 0.9827 & $-0.76$ \\
equidistant sources (EQ8)              & 0.9881 & $-0.22$ \\
random 8-channel pairing (RND8)        & 0.9870 & $-0.33$ \\
\emph{fusion} & & \\ 
fused source/target groups (FALL2)     & 0.9908 & $+0.06$ \\
\midrule
\multicolumn{3}{@{}l}{\emph{patch resolution (Real-IAD six-category subset)}} \\
overlapping patch-8, stride 8 (ours)   & 0.9350 & -- \\
native patch-14, stride 14             & 0.9016 & $-3.34$ \\
\bottomrule
\end{tabular}
\captionsetup{font=footnotesize}
\caption{Ablation on cable, screw, pcb1 and macaroni2 (mean image AUROC). Full channel sets and
per-category results are in the supplement. The patch-resolution block uses six Real-IAD
categories and is compared only within that subset.}
\label{tab:ablation}
\end{table}

Source identity has a smaller effect: equidistant and random sources stay within $0.33$~pp, one and
two deep sources lose $0.29$ and $0.10$~pp, and a fourth adds nothing. Target depth has a larger
effect: replacing L7 loses $2.76$~pp and a blind target loses $5.17$~pp. A supplementary one-target scan
changes half of the channels and shows a broad sensitive plateau; it supports a target segment,
not a uniquely optimal block. The branch groups are not interchangeable: forward-only loses
$2.39$~pp, while backward-only loses $0.28$~pp. Fusion gains at most $0.06$~pp.

Per-channel analysis on five categories shows complementary specialization. Cable's layout
changes are almost solved by $1{\to}14$ alone ($0.998$), screw is strongest through deep-to-L7
($21{\to}7$, $0.990$), and pcb1, macaroni2 and audiojack favor deep-to-L14 channels (best
$0.993$, $0.991$ and $0.849$). These are diagnostic per-channel measurements, not a rule for
selecting channels per category; the exact table is in the supplement. The model keeps the same
eight channels everywhere.

The complete ablation configurations and all measured results are provided in
Appendix~\ref{sec:suppl_ablation}.

\begin{table}[!t]
\centering
\footnotesize
\setlength{\tabcolsep}{3pt}
\begin{tabular}{l ccc}
\toprule
backbone & MVTec-AD I & VisA I & Real-IAD I/O \\
\midrule
ViT-L/14   & 99.75 & 99.42 & 94.27 / 96.26 \\
ViT-B/14   & 99.46 & 99.04 & 94.01 / 96.02 \\
RADIO-B    & 99.33 & 98.35 & 93.27 / 95.78 \\
\bottomrule
\end{tabular}
\caption{Backbone generalization at $448^2$ in image-level AUROC. The Real-IAD cell includes
object-level AUROC (I/O). Pixel columns are omitted because their protocol was refreshed only for
the final ViT-L run. The layer set changes only when the backbone changes.}
\label{tab:backbones}
\end{table}

\subsection{Generalization across Backbones}
\label{sec:general}

We repeat layer selection on 12-block DINOv2 ViT-B/14 and RADIO-B~\citep{ranzinger2024amradio}.
Both have a shallow notch and a late collapse, and the same rule yields $\{1,3,7,10,11\}$. Only the
layer indices change; the predictors, probe, scoring and floor rules remain the same.
Table~\ref{tab:backbones} reports the image-level comparison.

ViT-L remains strongest, ViT-B is close, and RADIO-B loses most on VisA. From ViT-L to ViT-B,
image AUROC drops by 0.29, 0.38 and 0.26~pp on MVTec-AD, VisA and Real-IAD. RADIO-B loses
1.07~pp image AUROC on VisA.

The change from ViT-L to ViT-B is smaller than the change from ViT-B to RADIO-B. On Real-IAD,
ViT-B keeps object AUROC within 0.24~pp of ViT-L. RADIO-B remains
competitive on MVTec-AD, but it loses most on VisA. Because RADIO mixes several pretraining
signals~\citep{ranzinger2024amradio}, we do not attribute this drop to one representation property.
These comparisons keep the
backbone fixed within each row and therefore isolate representation quality from layer selection.

\textbf{Cost and operating regime.} The backbone runs once and only the directed predictors and
foreground probes are trained. The eight predictors share the same token dimension and hidden
width, so inference consists of eight token-wise MLPs after one six-layer feature read. No test
image updates the model, and no channel is selected from test performance. We do not claim an
end-to-end efficiency ranking: wall-clock latency and peak memory were not measured under a
controlled hardware protocol. SAM-2 is needed only to build training pseudo-masks. When masks are
unavailable, the prior is disabled as a degraded fallback: removing it lowers image AUROC by
$0.14$, $1.05$ and $3.71$~pp on cable, pcb1 and macaroni2, respectively.

The mixed comparison with the Dinomaly series is useful for deployment choices. Our gains are
largest at the image level on Real-IAD and VisA, while the Dinomaly series remains stronger in
pixel ranking and on the nearly saturated MVTec-AD benchmark. Applications that need to decide
whether a part is abnormal can therefore use the image-level score; pixel localization remains the
weaker side of our method.

The main failure modes are also visible from the design. Multi-object scenes lose the global
branch because its mean mixes unrelated objects. Thin objects can produce incomplete SAM-2 masks,
which is why the reliability gate disables the prior when geometry or foreground/background
contrast is implausible. Finally, the method assumes that a category has a stable normal training
set; cross-category training would require recalibrating the background prior and the target-depth
statistics.

\subsection{Discussion}
\label{sec:discussion}

The fixed eight-channel set avoids label-dependent channel search and combines $1{\to}14$
reconfiguration, L7 fine structure and deep-to-L14 regional deviations. Using one or two deep
sources instead of three costs $0.29$ and $0.10$~pp, while leaving the sensitive segment costs over
$5$~pp. Target depth therefore matters more than source identity, which mainly provides a
robustness margin. Branch A corrects a per-image offset, and branch B removes a smooth spatial
background. The global branch captures whole-part shifts but cannot localise, so it is excluded
from the pixel metric; pixel scoring averages the eight residual maps directly.

The pointwise predictor cannot model cross-token information flow; its residual emphasises local
mismatch, consistent with the strength on scratches and texture. The Dinomaly family reconstructs layer groups
and is stronger in dense pixel ranking, whereas our score is stronger on image/object decisions.
The two methods are complementary; a coarse-to-fine combination remains untested. The
training-only foreground prior is gated and disabled when unreliable.

The main comparison uses direct square resizing. Its statistical limits remain: models are trained
independently per category, and main results are deterministic single runs, so we do not interpret
sub-percentage-point differences as stable category rankings.
The supplement reports category dispersion, but training-seed variance, calibration-budget
sensitivity and full cross-category training remain unestimated; the evidence therefore supports a
mechanism-level gain from the fixed channel set, rather than universal per-class superiority or a
claim that the calibration budgets are optimal.

\textbf{Single-class scope and multi-category extension.} We retain one model per category because the
benchmark defines a separate one-class normal support for each category and because the high-pass
strength, branch normalisation and foreground prior are estimated from category-specific normal
images; pooling categories without explicit conditioning would mix several normal manifolds, so a
change that is nominal for one object could become anomalous for another. Multi-category anomaly
detectors are therefore complementary~\citep{you2022uniad,guo2023recontrast,guo2025dinomaly,
he2024mambaad,zhang2025vitad}, but evaluating them requires a different protocol. The present choice
also exposes the main extension cost: adding a new single category requires training its eight
predictors and recalibrating the three category-specific statistics. A useful next test is a shared
predictor with category-conditioned calibration or lightweight mixture-of-experts heads, evaluated
on held-out-category and object-level protocols; full multi-category training and unseen-category
transfer remain outside the scope of this paper.

\section{Conclusion}
\label{sec:conclusion}

Adjacent-layer cosine similarity reveals stable depth segments. Using deep and shallow blocks as
sources and segment boundaries or midpoints as targets turns cross-layer transfer failure into an
anomaly score. The rule and scorer work with ViT-L, ViT-B and RADIO-B. Against the Dinomaly series,
ours is stronger on VisA/Real-IAD image AUROC; the series is stronger on MVTec image AUROC,
Real-IAD object AUROC and all pixel metrics under the refreshed protocol. The foreground prior is
training-only and gated. Pixel localization remains the main gap. A useful next test is to combine
the pointwise score with regional layer-group reconstruction.

\section*{LLM Usage Statement}

OpenAI Codex, DeepSeek, and Kimi were used to assist with English-language editing, manuscript organization,
LaTeX formatting, figure and table preparation, and code-assisted aggregation of experimental
results. The experimental data were produced by our implementation and verified by the authors.
All technical claims, numerical results, figures, and citations were checked against the source
code, logs, papers, and documentation. The authors take full responsibility for the content of
the submission.

\clearpage
\onecolumn
\appendix
\suppressfloats[t]
\section{Supplementary Material}
\label{sec:suppl}

The main $448^2$ table and per-category table use the refreshed direct-square, no-synthesis run.
The four-category ablations, backbone-transfer rows, high-pass tables and mask-gate counts are
matched mechanism studies from the preceding locked run. They are not used to compute absolute
deltas against the refreshed main scores.

\subsection{Extended Related Work}
\label{sec:suppl_related}

\textbf{Reconstruction family.} RD4AD~\citep{deng2022rd4ad} keeps a pretrained encoder frozen,
fuses multi-scale teacher features into a compact one-class embedding, and uses a student decoder
to reconstruct the teacher's multi-scale representations. DeSTSeg~\citep{zhang2023destseg} adds a
denoising student and a segmentation network to this skeleton.
The Dinomaly series~\citep{guo2025dinomaly,guo2025dinomaly2} aggregates eight ViT blocks into a
noisy MLP bottleneck and reconstructs grouped layer representations. Its
over-generalisation
argument is that strict layer-to-layer supervision can make the decoder too capable of restoring
anomalous content. Loose Reconstruction therefore groups multiple encoder layers into semantic
clusters, reconstructs the cluster sums, and randomly drops bottleneck activations.

This remedy changes how strongly the decoder is pushed to reproduce its input, but not what it
is asked to predict. A block that is intrinsically insensitive to a defect can therefore retain
a near-zero residual. In contrast, our deep-to-sensitive channels make a deep source predict a
sensitive target it cannot faithfully represent. Insensitivity then determines the anomaly signal instead of
suppressing it.

\textbf{Embedding and synthesis families.} PatchCore~\citep{roth2022patchcore} stores a coreset of
nominal patch features and uses nearest-neighbour distance, while PaDiM~\citep{defard2021padim}
models each spatial position with a Gaussian and scores by Mahalanobis distance. Synthetic-anomaly
methods include CutPaste~\citep{li2021cutpaste}, DRAEM~\citep{zavrtanik2021draem},
SimpleNet~\citep{liu2023simplenet} and GLASS~\citep{chen2024glass}; EfficientAD
distils a lightweight student from a teacher. Their performance is tied to the realism of the
synthetic defects or the capacity of the student, and none models directed relations along depth.

\textbf{Benchmarks and backbone context.} MVTec-AD~\citep{bergmann2019mvtec} contains 15
single-view categories, VisA~\citep{zou2022visa} contains more complex products with multiple
objects, and Real-IAD~\citep{wang2024realiad} contains 30 categories captured from five views.
MANTA~\citep{fan2025manta} targets tiny defects with high-resolution multi-view captures. We use
the per-category single-class protocol throughout and also report object-level Real-IAD
AUROC by grouping the five views of each object. The frozen DINOv2 backbone family supplies the
features on which the fault structure is measured.

\subsection{Per-Category High-Pass Calibration}
\label{sec:suppl_lambda}

Tables~\ref{tab:lambda-mvtec}--\ref{tab:lambda-riad} list the high-pass strength $\lambda$
produced by the training-only calibration described in the main paper's scoring section. Each value
follows directly from $\lambda=0.75\,\mathrm{clip}((\rho-0.80)/0.08,0,1)$ and the category's own
$\rho$. The filter switches off on 13 of 15 MVTec-AD categories, while
VisA pcb3 and Real-IAD switch reach the maximum of $0.75$.

\begin{table}[t]
\centering
\small
\begin{tabular}{lcc}
\toprule
category & $\lambda$ & $\rho$ \\
\midrule
bottle & 0.000 & 0.7960 \\
cable & 0.000 & 0.7123 \\
capsule & 0.000 & 0.6818 \\
carpet & 0.000 & 0.5752 \\
grid & 0.000 & 0.6649 \\
hazelnut & 0.000 & 0.7785 \\
leather & 0.000 & 0.6567 \\
metal\_nut & 0.000 & 0.7784 \\
pill & 0.000 & 0.7353 \\
screw & 0.000 & 0.7108 \\
tile & 0.000 & 0.5972 \\
toothbrush & 0.282 & 0.8301 \\
transistor & 0.000 & 0.6601 \\
wood & 0.000 & 0.6282 \\
zipper & 0.001 & 0.8001 \\
\bottomrule
\end{tabular}
\caption{MVTec-AD high-pass calibration.}
\label{tab:lambda-mvtec}
\end{table}

\begin{table}[t]
\centering
\small
\begin{tabular}{lcc}
\toprule
category & $\lambda$ & $\rho$ \\
\midrule
candle & 0.000 & 0.6455 \\
capsules & 0.631 & 0.8673 \\
cashew & 0.000 & 0.7130 \\
chewinggum & 0.000 & 0.7926 \\
fryum & 0.604 & 0.8644 \\
macaroni1 & 0.596 & 0.8635 \\
macaroni2 & 0.689 & 0.8735 \\
pcb1 & 0.199 & 0.8212 \\
pcb2 & 0.209 & 0.8223 \\
pcb3 & 0.750 & 0.9047 \\
pcb4 & 0.266 & 0.8284 \\
pipe\_fryum & 0.000 & 0.7461 \\
\bottomrule
\end{tabular}
\caption{VisA high-pass calibration.}
\label{tab:lambda-visa}
\end{table}

\begin{table}[t]
\centering
\small
\begin{tabular}{lcc}
\toprule
category & $\lambda$ & $\rho$ \\
\midrule
audiojack & 0.651 & 0.8695 \\
bottle\_cap & 0.000 & 0.7186 \\
button\_battery & 0.745 & 0.8795 \\
end\_cap & 0.530 & 0.8566 \\
eraser & 0.000 & 0.6983 \\
fire\_hood & 0.295 & 0.8315 \\
mint & 0.000 & 0.7479 \\
mounts & 0.000 & 0.7212 \\
pcb & 0.666 & 0.8710 \\
phone\_battery & 0.000 & 0.7546 \\
plastic\_nut & 0.000 & 0.7961 \\
plastic\_plug & 0.000 & 0.7693 \\
porcelain\_doll & 0.089 & 0.8095 \\
regulator & 0.000 & 0.7076 \\
rolled\_strip\_base & 0.000 & 0.6996 \\
sim\_card\_set & 0.000 & 0.6583 \\
switch & 0.750 & 0.8885 \\
tape & 0.000 & 0.7068 \\
terminalblock & 0.000 & 0.7178 \\
toothbrush & 0.066 & 0.8071 \\
toy & 0.000 & 0.7842 \\
toy\_brick & 0.125 & 0.8133 \\
transistor1 & 0.459 & 0.8490 \\
u\_block & 0.000 & 0.7732 \\
usb & 0.644 & 0.8686 \\
usb\_adaptor & 0.000 & 0.6860 \\
vcpill & 0.322 & 0.8344 \\
wooden\_beads & 0.502 & 0.8536 \\
woodstick & 0.154 & 0.8165 \\
zipper & 0.000 & 0.7934 \\
\bottomrule
\end{tabular}
\caption{Real-IAD high-pass calibration.}
\label{tab:lambda-riad}
\end{table}

\subsection{Calibration, Statistical Scope, and Comparison Boundaries}
\label{sec:suppl_calibration_scope}

\textbf{Sampling for training-only calibration.} All calibration quantities use normal training
images only. The A/B/global z-score uses the first 60 images in deterministic loader order. The
high-pass coefficient uses the first 25 normal training images, while the mask-gate statistics use
the first 25 cached masks. The foreground probe uses up to 50 cached masks, samples 800 patches per
image with a fixed seed, and fits an L2 logistic regression with $C=1$. The mask-generation stage
samples at most 50 images with fixed seed 555, so production runs use at most 50 masks per category.
These caps keep calibration cost fixed for Real-IAD categories that contain more than 1,000 normal
images. We did not perform a sensitivity study over the 25/50/60/800 sample budgets, so they should
be read as fixed implementation choices rather than tuned optima.

\textbf{Thresholds and category adaptation.} The fault detector, channel set, MLP architecture,
top-0.2\% pooling, floor values and gate thresholds are shared across datasets. Three quantities
adapt to a category: $\lambda$ is a differentiable function of its training residual spectrum; the
branch statistics are normalised on its training images; and the mask prior is fitted from its
training pseudo-masks. The floor threshold $BC=0.40$ separates observed training-only values of
0.266 and 0.421, and the reliability-gate constants were then held across all 57 main categories.
We also swept the BC threshold over $[0.2,0.6]$ using the paired $f=0.1$ and $f=0.7$
predictions available for the 27 mask-on/off categories. The MVTec mean changes by less than
0.01~pp over this range, and the VisA mean varies by at most 0.056~pp. The choice is therefore
operationally stable on this diagnostic set, although it should not be interpreted as a formal
held-out optimality result: a high $BC$ can indicate either background distractors or an
incomplete mask, and the reliability gate is the primary safeguard for the latter.

\textbf{Category dispersion.} Table~\ref{tab:category-dispersion} reports the mean and standard
deviation over categories for the refreshed production ViT-L/14 results. These values describe
heterogeneity across object categories, not variance over repeated training seeds. In particular,
the Real-IAD image-level standard deviation is $3.51$~pp across 30 categories, so its aggregate
gap should not
be interpreted as a uniform improvement on every object.

\begin{table}[t]
\centering
\small
\setlength{\tabcolsep}{3pt}
\begin{tabular}{@{}lcccc@{}}
\toprule
dataset & I-AUROC & O-AUROC & P-AUROC & P-AUPRO \\
\midrule
MVTec-AD & $99.75\pm0.34$ & -- & $98.46\pm1.22$ & $95.97\pm4.12$ \\
VisA     & $99.42\pm0.53$ & -- & $99.21\pm0.94$ & $96.12\pm1.98$ \\
Real-IAD & $94.27\pm3.51$ & $96.26\pm2.89$ & $99.26\pm1.12$ & $96.59\pm3.76$ \\
\bottomrule
\end{tabular}
\caption{Category-level mean $\pm$ standard deviation for the production ViT-L/14 model. Standard
deviations are computed over categories, not repeated runs.}\label{tab:category-dispersion}
\end{table}

\textbf{Baseline and mechanism controls not completed.} The reference Real-IAD table is retained
for benchmark calibration, not as a protocol-identical ranking. We did not reproduce all recent
single-class detectors or run a separate layer-group reconstruction model with our complete scoring
pipeline. Such a control is necessary to quantify how much of the result is attributable to the
cross-layer residual versus the shared post-processing. Our existing patch-resolution, direction,
target-depth and source-count controls isolate parts of this question, but they do not replace that
experiment. We also report no repeated-seed variance, wall-clock efficiency, F1 at a fixed operating
threshold, or a qualitative failure gallery. These omissions define the quantitative scope of the
current submission.

\textbf{Ablation breadth and loss weighting.} The controlled component ablation uses cable, screw,
pcb1 and macaroni2; the extra classes in the per-channel discussion are diagnostic examples, not
part of the controlled mean. We did not repeat every ablation configuration across all 57 main
categories. The training objective averages the two forward channels and six backward channels
within their respective directions, then weights the two direction means equally. This is
intentional branch balancing, but it gives each forward channel three times the per-channel weight
of a backward channel; the forward-only and backward-only rows quantify the resulting trade-off.

\subsection{Complete Ablation Results}
\label{sec:suppl_ablation}

Table~\ref{tab:ablation-configs} gives the exact channel set behind every abbreviation used in the
main ablation. BWD4 is the historical four-channel subset obtained by dropping source block $23$;
BWD6 contains all six backward channels. A missing entry denotes that the corresponding branch is
disabled.

\begin{longtable}{@{}p{0.17\textwidth}p{0.38\textwidth}p{0.38\textwidth}@{}}
\caption{Exact source-to-target sets for the reported ablation configurations.}\label{tab:ablation-configs}\\
\toprule
configuration & forward channels & backward channels \\
\midrule
\endfirsthead
\toprule
configuration & forward channels & backward channels \\
\midrule
\endhead
\bottomrule
\endlastfoot
Ours (8) &
$f_{1\to14},f_{1\to7}$ &
$b_{21\to14},b_{21\to7},b_{22\to14},b_{22\to7},b_{23\to14},b_{23\to7}$ \\
forward-only &
$f_{1\to14},f_{1\to7}$ & -- \\
BWD4 &
-- &
$b_{21\to14},b_{21\to7},b_{22\to14},b_{22\to7}$ \\
BWD6 &
-- &
$b_{21\to14},b_{21\to7},b_{22\to14},b_{22\to7},b_{23\to14},b_{23\to7}$ \\
B1 &
$f_{1\to14},f_{1\to7}$ &
$b_{23\to14},b_{23\to7}$ \\
B2 &
$f_{1\to14},f_{1\to7}$ &
$b_{22\to14},b_{22\to7},b_{23\to14},b_{23\to7}$ \\
B3 (= Ours) &
$f_{1\to14},f_{1\to7}$ &
$b_{21\to14},b_{21\to7},b_{22\to14},b_{22\to7},b_{23\to14},b_{23\to7}$ \\
B4 &
$f_{1\to14},f_{1\to7}$ &
$b_{20\to14},b_{20\to7},b_{21\to14},b_{21\to7},$ \newline
$b_{22\to14},b_{22\to7},b_{23\to14},b_{23\to7}$ \\
B7 &
$f_{14\to7}$ & $b_{14\to7}$ \\
B8 &
$f_{14\to8}$ & $b_{14\to8}$ \\
ONLY14 &
$f_{1\to14}$ &
$b_{21\to14},b_{22\to14},b_{23\to14}$ \\
EQ8 &
$f_{2\to14},f_{2\to7},f_{9\to14},f_{9\to7},f_{16\to14},f_{16\to7}$ &
$b_{23\to14},b_{23\to7}$ \\
RND8 &
$f_{2\to16},f_{9\to17},f_{13\to16},f_{14\to17}$ &
$b_{18\to9},b_{19\to11},b_{20\to13},b_{20\to15}$ \\
FAR4 &
$f_{1\to22},f_{2\to23}$ &
$b_{22\to1},b_{23\to2}$ \\
deep-source control &
$f_{1\to14},f_{1\to7}$ &
$b_{8\to14},b_{8\to7},b_{9\to14},b_{9\to7},b_{10\to14},b_{10\to7}$ \\
TGT0 &
$f_{1\to14},f_{1\to0}$ &
$b_{21\to14},b_{21\to0},b_{22\to14},b_{22\to0},b_{23\to14},b_{23\to0}$ \\
FALL2 &
$f_{\{0,1\}\to\{12,14,16\}}$, $f_{\{0,1\}\to\{7,8\}}$ &
$b_{\{21,22\}\to\{12,14,16\}}$, $b_{\{21,22\}\to\{7,8\}}$,
$b_{\{22,23\}\to\{12,14,16\}}$, $b_{\{22,23\}\to\{7,8\}}$ \\
MEAN / MEDIAN &
\multicolumn{2}{l}{Same eight production channels; only score aggregation changes.} \\
\end{longtable}

Table~\ref{tab:ablation-full} reports every measured configuration on the four fixed
categories. All rows use the production soft-gated scoring and differ from the matched
reference in exactly one design variable. Table~\ref{tab:source-count-detail} further reports the
per-category results for the backward-direction and source-count subsets.

\begin{longtable}{llrr}
\caption{Complete four-category ablation results. I-ROC is reported as a fraction; $\Delta$
is relative to the matched no-input-injection reference (NOINP).}\label{tab:ablation-full}\\
\toprule
group & configuration & mean I-ROC & $\Delta$ (pp) \\
\midrule
\endfirsthead
\toprule
group & configuration & mean I-ROC & $\Delta$ (pp) \\
\midrule
\endhead
\bottomrule
\endlastfoot
reference & NOINP & 0.9902 & -- \\
control & F8NOINP & 0.9901 & $-0.02$ \\
\midrule
direction & forward-only & 0.9663 & $-2.39$ \\
          & BWD4 & 0.9868 & $-0.34$ \\
          & BWD6 & 0.9874 & $-0.28$ \\
\midrule
source & A0 & 0.9901 & $-0.01$ \\
       & B1 & 0.9873 & $-0.29$ \\
       & B2 & 0.9892 & $-0.10$ \\
       & B4 & 0.9902 & $0.00$ \\
       & E1 & 0.9903 & $+0.01$ \\
       & EQ8 & 0.9881 & $-0.22$ \\
       & deep-source control & 0.9827 & $-0.76$ \\
       & SRCS2B & 0.9865 & $-0.37$ \\
       & RND8 & 0.9870 & $-0.33$ \\
\midrule
target & A1 & 0.9893 & $-0.10$ \\
       & A2 & 0.9895 & $-0.07$ \\
       & B7 & 0.9626 & $-2.76$ \\
       & B8 & 0.9676 & $-2.27$ \\
       & T10 & 0.9885 & $-0.17$ \\
       & T12 & 0.9894 & $-0.08$ \\
       & T16 & 0.9899 & $-0.04$ \\
       & T18 & 0.9860 & $-0.43$ \\
       & TNear3 & 0.9880 & $-0.22$ \\
       & TMid18 & 0.9845 & $-0.57$ \\
       & TGT0 & 0.9795 & $-1.07$ \\
       & TGT2 & 0.9881 & $-0.21$ \\
       & FAR4 & 0.9386 & $-5.17$ \\
       & ONLY14 & 0.9866 & $-0.36$ \\
\midrule
fusion & F0 & 0.9901 & $-0.01$ \\
       & F01 & 0.9899 & $-0.03$ \\
       & F78 & 0.9901 & $-0.01$ \\
       & F8 & 0.9896 & $-0.06$ \\
       & F1416 & 0.9900 & $-0.02$ \\
       & F2122 & 0.9900 & $-0.02$ \\
       & F2123 & 0.9882 & $-0.20$ \\
       & F2223 & 0.9899 & $-0.03$ \\
       & FALL & 0.9888 & $-0.14$ \\
       & FALL2 & 0.9908 & $+0.06$ \\
\end{longtable}

\begin{table}[t]
\centering
\scriptsize
\setlength{\tabcolsep}{2pt}
\begin{tabular}{@{}lccccr@{}}
\toprule
configuration & cable & screw & pcb1 & macaroni2 & mean \\
\midrule
BWD6             & 0.9888 & 0.9904 & 0.9969 & 0.9733 & 0.9874 \\
B1 (1 deep source)  & 0.9981 & 0.9873 & 0.9927 & 0.9711 & 0.9873 \\
B2 (2 deep sources) & 0.9972 & 0.9902 & 0.9970 & 0.9722 & 0.9892 \\
B4 (4 deep sources) & 0.9951 & 0.9918 & 0.9980 & 0.9758 & 0.9902 \\
\bottomrule
\end{tabular}
\caption{Per-category results for the new direction and deep-source-count ablations. B3 is the
production reference (three deep sources) with mean $0.9902$. Adding a fourth source matches B3,
so accuracy saturates at the production set.}
\label{tab:source-count-detail}
\end{table}

\subsection{Per-Channel Specialization}
\label{sec:suppl_channel_roles}

Table~\ref{tab:channel-roles} reports image-level AUROC when each directed channel is scored
alone. The five categories cover structural layout changes (cable), fine geometry (screw), circuit
patterns (pcb1), multi-object texture (macaroni2), and a small mixed-domain object (audiojack).
The best channel changes with the anomaly type while the eight-channel set stays fixed.

\begin{table}[t]
\centering
\scriptsize
\setlength{\tabcolsep}{2pt}
\begin{tabular}{@{}lccccc@{}}
\toprule
category & $1{\to}14$ & $1{\to}7$ & best deep$\to$14 & best deep$\to$7 & mean \\
\midrule
cable     & \textbf{0.9981} & 0.9875 & 0.9893 ({23}$\to$14) & \textbf{0.9920} ({21}$\to$7) & 0.9985 \\
screw     & 0.9283 & 0.9524 & 0.9889 ({22}$\to$14) & \textbf{0.9902} ({21}$\to$7) & 0.9920 \\
pcb1      & 0.9818 & 0.9889 & \textbf{0.9927} ({23}$\to$14) & 0.9862 ({23}$\to$7) & 0.9918 \\
macaroni2 & 0.9286 & 0.9811 & \textbf{0.9907} ({23}$\to$14) & 0.9893 ({23}$\to$7) & 0.9915 \\
audiojack & 0.7852 & 0.8358 & \textbf{0.8485} ({23}$\to$14) & 0.8393 ({21}$\to$7) & 0.8486 \\
\bottomrule
\end{tabular}
\caption{Per-channel image AUROC on five diagnostic categories. ``Best'' is selected only for
display within each category; the deployed model always uses all eight channels. The table
illustrates channel specialization and is not used for per-category selection. The final mean
column evaluates all eight channels together.}
\label{tab:channel-roles}
\end{table}

The winning channel depends on the anomaly type rather than depth alone. Cable is nearly solved by the forward
$1{\to}14$ path, whereas screw benefits most from deep information returned to the near target
L7. The circuit and multi-object categories instead favor deep-to-L14 mappings. Audiojack is the
hardest of the five and also peaks at $23{\to}14$. This variation is the empirical basis for
retaining a fixed set of roles instead of choosing one layer pair for every category.

\subsection{Foreground-Prior Ablation}
\label{sec:suppl_prior_ablation}

Table~\ref{tab:prior-ablation} compares the production foreground weighting with the same score
when $w_p\equiv1$. The checkpoint, test images, eight channels and post-processing remain fixed;
only the foreground weight changes. A category whose reliability gate disables the mask has
identical values in the two columns by construction.

\begin{table}[t]
\centering
\small
\setlength{\tabcolsep}{4pt}
\begin{tabular}{@{}lccrr@{}}
\toprule
category & gate & prior off & prior on & $\Delta$ (pp) \\
\midrule
cable     & ON  & 0.9971 & 0.9985 & $+0.14$ \\
screw     & OFF & 0.9920 & 0.9920 & $0.00$ \\
pcb1      & ON  & 0.9813 & 0.9918 & $+1.05$ \\
macaroni2 & ON  & 0.9544 & 0.9915 & $+3.71$ \\
\bottomrule
\end{tabular}
\caption{Image-level AUROC with and without the foreground prior. The prior is required for
accepted masks and can be decisive when background evidence competes with a defect. The reliability
gate keeps it disabled when the training mask is unreliable.}
\label{tab:prior-ablation}
\end{table}

The effect is not uniform: cable changes by only $0.14$~pp, while pcb1 and macaroni2 gain
$1.05$ and $3.71$~pp. The foreground prior is therefore a structural component of the
pointwise-residual detector, not optional post-processing. The gate determines when the component
is safe to apply.

\subsection{Mechanism-Control Pilot}
\label{sec:suppl_mechanism_pilot}

Table~\ref{tab:mechanism-pilot} compares the cross-layer $21{\to}14$ map with two alternative
objectives under the same normal tokens, training steps, foreground gate and post-processing. The
ordinary reconstruction head
reconstructs $f_{14}$ through a bottleneck; the grouped head reconstructs $f_{14}$ from the mean
of $f_7$ and $f_{14}$. All three heads have 1.58M parameters.

\begin{table}[t]
\centering
\small
\setlength{\tabcolsep}{4pt}
\begin{tabular}{@{}lccc@{}}
\toprule
category & cross $21{\to}14$ & recon $14$ & grouped recon \\
\midrule
cable     & \textbf{0.9648} & 0.9565 & 0.9515 \\
screw     & \textbf{0.9553} & 0.9076 & 0.9129 \\
pcb1      & \textbf{0.9658} & 0.9602 & 0.9630 \\
macaroni2 & 0.9637 & 0.9806 & \textbf{0.9813} \\
\bottomrule
\end{tabular}
\caption{Mechanism-control pilot under matched head capacity and identical scoring. Cross-layer
prediction wins on cable, screw and pcb1, while reconstruction is stronger on macaroni2. This is a
four-category local pilot, not the full six-category study, and it motivates a complementary rather
than universal claim.}
\label{tab:mechanism-pilot}
\end{table}

Cross-layer prediction is useful where fine local evidence is difficult to transport, but the
results do not support universal superiority over reconstruction. A Real-IAD extension is still
required before making a dataset-level claim.

\subsection{Marginal Target-Layer Sensitivity on VisA}
\label{sec:suppl_target_scan}

This experiment measures the marginal sensitivity of one target group in the complete eight-channel
model; it does not rank target blocks in isolation. Scan A replaces the four channels targeting
$L_7$ with channels targeting $t$, while retaining the four channels targeting $L_{14}$. Scan B
performs the converse: it replaces the four $L_{14}$ channels and retains the four $L_7$ channels.
Each point therefore changes exactly half of the predictor set, with training settings fixed. We
evaluate VisA pcb1 and macaroni2, the two categories used for layer analysis. The curves show mean
image AUROC and the bands span the two category values; Fig.~\ref{fig:suppl_target_scan} reports
the resulting curves.

\begin{center}
\includegraphics[width=0.64\linewidth]{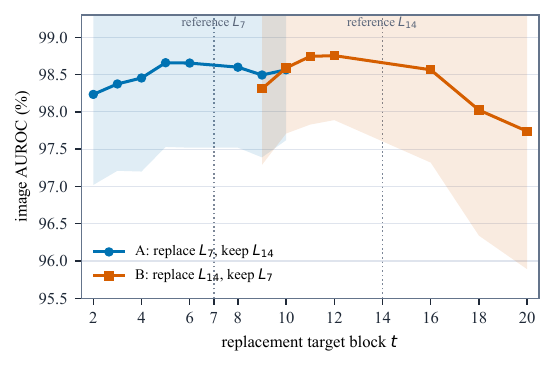}
\captionof{figure}{Marginal target sensitivity on VisA. Each scan replaces four channels associated
with one target while retaining the four channels associated with the other. Dotted lines mark the
production targets $L_7$ and $L_{14}$. The broad plateaus show robustness inside the sensitive
segment; they are not evidence that $L_7$ or $L_{14}$ is a uniquely optimal block. The $t=13$ point
is omitted because its checkpoint is unavailable under the locked protocol.}
\label{fig:suppl_target_scan}
\end{center}

Scan A varies from $98.24$ to $98.66$ and stays essentially tied with the fixed $L_7/L_{14}$
reference at $98.67$. Scan B peaks at $98.76$ around blocks 11--12 and then declines; moving to
blocks 18 and 20 loses $0.64$ and $0.93$ percentage points relative to the reference. The plateau
is partly expected because the untouched target group still contributes four channels, so the
figure should be read as a marginal-substitution test. Used this way, it supports two bounded
claims. First, the full predictor set does not require a category-specific search for a unique
target block. Second, the target still has to remain in the sensitive depth segment: the broad
plateau disappears near the late collapse, consistent with the larger loss of the blind-target
control in the main text.

\subsection{Patch-Resolution Control}
\label{sec:suppl_patch_resolution}

Table~\ref{tab:patch-resolution} reports the matched per-category and mean results.
The patch-resolution control removes the overlapping stride-8 patchization used by the production
model and instead uses the native DINOv2 $14\times14$ patch embedding with stride $14$. At
$448^2$ this produces a $32\times32$ token grid instead of $56\times56$. We retain the six selected
transformer blocks, the same eight directed source--target channels, the MLP predictor and the
same scoring pipeline; the mask probe and floor calibration are re-fitted on the native token
grid. Because this control was run on six Real-IAD categories, we compare it only with the
production model on the identical six-category subset.

\begin{table}[H]
\centering
\footnotesize
\setlength{\tabcolsep}{3pt}
\begin{tabular}{@{}lccc@{}}
\toprule
category & patch-8 & native-14 & $\Delta$ (pp) \\
\midrule
bottle\_cap     & 97.84 & 93.54 & $-4.30$ \\
button\_battery & 93.83 & 88.46 & $-5.37$ \\
end\_cap        & 91.89 & 88.17 & $-3.72$ \\
eraser          & 96.31 & 92.93 & $-3.38$ \\
fire\_hood      & 93.48 & 91.71 & $-1.77$ \\
mounts          & 87.63 & 86.13 & $-1.50$ \\
\midrule
mean            & 93.50 & 90.16 & $-3.34$ \\
\bottomrule
\end{tabular}
\caption{Image-level AUROC for the patch-resolution control on six Real-IAD categories.
\texttt{patch-8} is the overlapping stride-8 tokenization; \texttt{native-14} uses DINOv2's native patch
embedding. Removing the overlapping patchization lowers the matched mean by $3.34$~pp.}
\label{tab:patch-resolution}
\end{table}

\subsection{Additional Dataset Results}
\label{sec:suppl_extra_datasets}

Table~\ref{tab:extra-datasets} reports the four additional datasets evaluated with the same
ViT-L/14, $448^2$, MLP, mask-head and scoring protocol: MPDD~\citep{jezek2021mpdd},
BTAD~\citep{mishra2021btad}, DAGM~\citep{wieler2007dagm}, and Kolektor Surface-Defect
Dataset~2 (KSDD2)~\citep{bozic2021ksdd2}.
The official BTAD-02 normal training split contains unidentified anomalies; we retain it for
comparability~\citep{jiang2022softpatch,guo2025dinomaly2}. For DAGM, files listed under \texttt{Train/Label} are
excluded from normal training. KSDD2 has no ground-truth anomaly masks in the current release, so
we report only image-level AUROC and mark the pixel columns unavailable when ground truth is absent.

These results are a portability check, not a matched-baseline ranking: we did not retrain baselines
for BTAD, DAGM or KSDD2 and therefore do not compare quoted literature numbers across different
protocols. MPDD has a matched run from the Dinomaly series: the refreshed model leads by $0.45$~pp
in image AUROC but trails by $1.08$~pp in P-AUROC and $1.38$~pp in P-AUPRO (ours:
$98.27/98.24/95.73$; Dinomaly series: $97.82/99.32/97.11$). We use the additional datasets as a
portability check, not as evidence for a universal accuracy advantage.

\begin{longtable}{llrrr}
\caption{Per-category results on the additional datasets. All values are percentages; pixel
metrics use the same $256\times256$ evaluation grid as the refreshed main experiments.}\label{tab:extra-datasets}\\
\toprule
dataset & category & I-AUROC & P-AUROC & P-AUPRO \\
\midrule
\endfirsthead
\toprule
dataset & category & I-AUROC & P-AUROC & P-AUPRO \\
\midrule
\endhead
\bottomrule
\endlastfoot
MPDD & bracket\_black & 92.75 & 99.16 & 98.42 \\
     & bracket\_brown & 97.13 & 94.26 & 93.13 \\
     & bracket\_white & 99.78 & 99.59 & 94.86 \\
     & connector & 100.00 & 97.82 & 93.17 \\
     & metal\_plate & 100.00 & 99.53 & 97.98 \\
     & tubes & 99.95 & 99.10 & 96.82 \\
\midrule
BTAD & 01 & 100.00 & 98.52 & 89.78 \\
     & 02 & 88.80 & 97.22 & 61.56 \\
     & 03 & 99.59 & 99.91 & 99.42 \\
\midrule
DAGM & Class1 & 100.00 & 89.69 & 81.81 \\
     & Class2 & 100.00 & 99.71 & 99.69 \\
     & Class3 & 100.00 & 95.93 & 94.27 \\
     & Class4 & 100.00 & 99.42 & 98.49 \\
     & Class5 & 100.00 & 97.97 & 95.91 \\
     & Class6 & 100.00 & 99.16 & 98.20 \\
     & Class7 & 100.00 & 96.92 & 95.30 \\
     & Class8 & 100.00 & 99.16 & 99.34 \\
     & Class9 & 100.00 & 99.99 & 99.96 \\
     & Class10 & 100.00 & 99.94 & 99.85 \\
\midrule
KSDD2 & ksdd2 & 97.13 & -- & -- \\
\midrule
\multicolumn{2}{r}{MPDD mean} & 98.27 & 98.24 & 95.73 \\
\multicolumn{2}{r}{BTAD mean} & 96.13 & 98.55 & 83.59 \\
\multicolumn{2}{r}{DAGM mean} & 100.00 & 97.79 & 96.28 \\
\multicolumn{2}{r}{KSDD2} & 97.13 & -- & -- \\
\end{longtable}

\subsection{Mask-Gate and Floor Decisions}
\label{sec:suppl_mask_gate}

Table~\ref{tab:mask-gate-decisions} lists the deployment decisions from the training-only mask
gate. The floor column counts categories ultimately assigned $0.1$ versus $0.7$; categories
without enough SAM2 masks do not receive a foreground prior.

\begin{table}[H]
\centering
\scriptsize
\setlength{\tabcolsep}{2pt}
\begin{tabular}{@{}lrrrrr@{}}
\toprule
dataset & categories & mask ON & gate OFF & no mask & $f=0.1$ \\
\midrule
MVTec-AD & 15 & 7 & 3 & 5 & 2 \\
VisA & 12 & 11 & 1 & 0 & 9 \\
Real-IAD & 30 & 30 & 0 & 0 & 2 \\
\bottomrule
\end{tabular}
\caption{Mask-gate and floor assignments for the production ViT-L/14 runs. Gate-off categories
are capsule, screw and transistor on MVTec-AD, and pcb3 on VisA. Real-IAD is gate-on for all 30
categories.}\label{tab:mask-gate-decisions}
\end{table}

\subsection{Per-Category Production Results}
\label{sec:suppl_categories}

Table~\ref{tab:per-category} reports the category-level production results.
{\footnotesize
\begin{longtable}{llccc}
\caption{Per-category ViT-L/14 results for the refreshed $448^2$ run. I, O, P and PRO denote image AUROC, object AUROC, P-AUROC and P-AUPRO.}
\label{tab:per-category}\\
\toprule
dataset & category & I (O) & P & PRO \\
\midrule
\endfirsthead
\toprule
dataset & category & I (O) & P & PRO \\
\midrule
\endhead
\bottomrule
\endlastfoot
MVTec-AD & \texttt{bottle} & 100.00 & 99.20 & 97.70 \\
 & \texttt{cable} & 99.55 & 98.69 & 95.79 \\
 & \texttt{capsule} & 99.16 & 98.79 & 97.69 \\
 & \texttt{carpet} & 100.00 & 99.43 & 98.35 \\
 & \texttt{grid} & 100.00 & 99.44 & 96.78 \\
 & \texttt{hazelnut} & 100.00 & 99.55 & 98.11 \\
 & \texttt{leather} & 100.00 & 99.37 & 98.68 \\
 & \texttt{metal\_nut} & 100.00 & 98.48 & 96.68 \\
 & \texttt{pill} & 99.67 & 96.86 & 97.94 \\
 & \texttt{screw} & 99.04 & 99.45 & 97.15 \\
 & \texttt{tile} & 100.00 & 97.07 & 92.72 \\
 & \texttt{toothbrush} & 100.00 & 98.97 & 96.82 \\
 & \texttt{transistor} & 99.33 & 95.63 & 82.00 \\
 & \texttt{wood} & 99.65 & 97.01 & 96.40 \\
 & \texttt{zipper} & 99.84 & 98.96 & 96.79 \\
\midrule
VisA & \texttt{candle} & 99.05 & 99.66 & 97.35 \\
 & \texttt{capsules} & 99.98 & 99.48 & 98.52 \\
 & \texttt{cashew} & 99.76 & 99.06 & 95.79 \\
 & \texttt{chewinggum} & 99.96 & 99.49 & 93.31 \\
 & \texttt{fryum} & 99.56 & 96.35 & 92.52 \\
 & \texttt{macaroni1} & 100.00 & 99.81 & 97.12 \\
 & \texttt{macaroni2} & 98.60 & 99.86 & 98.75 \\
 & \texttt{pcb1} & 99.34 & 99.73 & 97.42 \\
 & \texttt{pcb2} & 99.01 & 99.45 & 94.81 \\
 & \texttt{pcb3} & 99.11 & 99.49 & 95.59 \\
 & \texttt{pcb4} & 98.62 & 98.98 & 94.86 \\
 & \texttt{pipe\_fryum} & 100.00 & 99.20 & 97.36 \\
\midrule
Real-IAD & \texttt{audiojack} & 94.50 (94.28) & 99.74 & 97.92 \\
 & \texttt{bottle\_cap} & 97.99 (98.84) & 99.89 & 98.84 \\
 & \texttt{button\_battery} & 94.10 (95.26) & 99.39 & 96.05 \\
 & \texttt{end\_cap} & 91.85 (95.74) & 99.69 & 98.78 \\
 & \texttt{eraser} & 96.38 (96.87) & 99.86 & 99.15 \\
 & \texttt{fire\_hood} & 93.93 (95.87) & 99.73 & 96.91 \\
 & \texttt{mint} & 85.98 (89.63) & 98.90 & 91.27 \\
 & \texttt{mounts} & 87.78 (99.30) & 99.66 & 97.33 \\
 & \texttt{pcb} & 93.49 (93.97) & 99.64 & 98.23 \\
 & \texttt{phone\_battery} & 95.35 (95.20) & 99.83 & 98.68 \\
 & \texttt{plastic\_nut} & 95.33 (98.56) & 99.89 & 99.46 \\
 & \texttt{plastic\_plug} & 93.83 (96.73) & 99.26 & 96.57 \\
 & \texttt{porcelain\_doll} & 92.55 (96.35) & 99.08 & 95.77 \\
 & \texttt{regulator} & 92.12 (94.90) & 99.74 & 98.02 \\
 & \texttt{rolled\_strip\_base} & 99.53 (99.43) & 99.86 & 99.31 \\
 & \texttt{sim\_card\_set} & 96.56 (96.97) & 99.74 & 97.52 \\
 & \texttt{switch} & 98.65 (99.33) & 98.87 & 98.68 \\
 & \texttt{tape} & 98.65 (99.88) & 99.88 & 99.10 \\
 & \texttt{terminalblock} & 98.71 (98.77) & 99.86 & 99.50 \\
 & \texttt{toothbrush} & 87.76 (97.30) & 95.04 & 81.48 \\
 & \texttt{toy} & 90.96 (91.42) & 95.68 & 93.05 \\
 & \texttt{toy\_brick} & 90.54 (91.51) & 98.67 & 89.18 \\
 & \texttt{transistor1} & 96.92 (99.14) & 99.55 & 98.20 \\
 & \texttt{u\_block} & 95.73 (98.63) & 99.75 & 98.22 \\
 & \texttt{usb} & 95.12 (96.26) & 99.26 & 98.40 \\
 & \texttt{usb\_adaptor} & 91.71 (94.74) & 99.62 & 97.34 \\
 & \texttt{vcpill} & 95.96 (99.06) & 99.22 & 93.42 \\
 & \texttt{wooden\_beads} & 94.11 (92.70) & 99.66 & 96.14 \\
 & \texttt{woodstick} & 92.18 (91.23) & 99.67 & 96.88 \\
 & \texttt{zipper} & 99.70 (99.97) & 99.24 & 98.19 \\
\end{longtable}
}

\clearpage
{
    \small
    \bibliographystyle{ieeenat_fullname}
    \bibliography{main}

\begin{thebibliography}{42}
\providecommand{\natexlab}[1]{#1}
\providecommand{\url}[1]{\texttt{#1}}
\expandafter\ifx\csname urlstyle\endcsname\relax
  \providecommand{\doi}[1]{doi: #1}\else
  \providecommand{\doi}{doi: \begingroup \urlstyle{rm}\Url}\fi

\bibitem[Aswani and Jabari(2026)]{aswani2026dmdvit}
Nishant~Suresh Aswani and Saif~Eddin Jabari.
\newblock Dynamic mode decomposition along depth in vision transformers.
\newblock arXiv preprint arXiv:2605.07556, 2026.

\bibitem[Batzner et~al.(2024)Batzner, Heckler, and
  K{\"o}nig]{batzner2024efficientad}
Kilian Batzner, Lars Heckler, and Rebecca K{\"o}nig.
\newblock {EfficientAD}: Accurate visual anomaly detection at millisecond-level
  latencies.
\newblock In \emph{IEEE Winter Conf. Appl. Comput. Vis. (WACV)}, pages
  127--137, 2024.

\bibitem[Bergmann et~al.(2019)Bergmann, Fauser, Sattlegger, and
  Steger]{bergmann2019mvtec}
Paul Bergmann, Michael Fauser, David Sattlegger, and Carsten Steger.
\newblock {MVTec AD} -- a comprehensive real-world dataset for unsupervised
  anomaly detection.
\newblock In \emph{CVPR}, pages 9592--9600, 2019.

\bibitem[Bo{\v z}i{\v c} et~al.(2021)Bo{\v z}i{\v c}, Tabernik, and Sko{\v
  c}aj]{bozic2021ksdd2}
Jakob Bo{\v z}i{\v c}, Domen Tabernik, and Danijel Sko{\v c}aj.
\newblock Mixed supervision for surface-defect detection: From weakly to fully
  supervised learning.
\newblock \emph{Computers in Industry}, 129:\penalty0 103459, 2021.

\bibitem[Chen et~al.(2024)Chen, Luo, Lv, and Zhang]{chen2024glass}
Qiyu Chen, Huiyuan Luo, Chengkan Lv, and Zhengtao Zhang.
\newblock A unified anomaly synthesis strategy with gradient ascent for
  industrial anomaly detection and localization.
\newblock In \emph{ECCV}, pages 37--54, 2024.

\bibitem[Chen et~al.(2018)Chen, Rubanova, Bettencourt, and
  Duvenaud]{chen2018neuralode}
Ricky T.~Q. Chen, Yulia Rubanova, Jesse Bettencourt, and David~K. Duvenaud.
\newblock Neural ordinary differential equations.
\newblock In \emph{Advances in Neural Information Processing Systems}, 2018.

\bibitem[Colbrook and Townsend(2024)]{colbrook2024rigorous}
Matthew~J. Colbrook and Alex Townsend.
\newblock Rigorous data-driven computation of spectral properties of koopman
  operators for dynamical systems.
\newblock \emph{Communications on Pure and Applied Mathematics}, 77\penalty0
  (1):\penalty0 221--283, 2024.

\bibitem[Costanzino et~al.(2025)Costanzino, Ramirez, Lisanti, and
  Stefano]{costanzino2025tinydefects}
Alex Costanzino, Pierluigi~Zama Ramirez, Giuseppe Lisanti, and Luigi~Di
  Stefano.
\newblock Learning to be a transformer to pinpoint anomalies.
\newblock \emph{IEEE Access}, 13:\penalty0 109624--109637, 2025.
\newblock arXiv:2407.04092.

\bibitem[Darcet et~al.(2024)Darcet, Oquab, Mairal, and
  Bojanowski]{darcet2024registers}
Timoth{\'e}e Darcet, Maxime Oquab, Julien Mairal, and Piotr Bojanowski.
\newblock Vision transformers need registers.
\newblock In \emph{ICLR}, 2024.

\bibitem[Defard et~al.(2021)Defard, Setkov, Loesch, and
  Audigier]{defard2021padim}
Thomas Defard, Aleksandr Setkov, Angelique Loesch, and Romaric Audigier.
\newblock {PaDiM}: A patch distribution modeling framework for anomaly
  detection and localization.
\newblock In \emph{ICPR}, pages 475--489, 2021.

\bibitem[Deng and Li(2022)]{deng2022rd4ad}
Hanqiu Deng and Xingyu Li.
\newblock Anomaly detection via reverse distillation from one-class embedding.
\newblock In \emph{CVPR}, pages 9737--9746, 2022.

\bibitem[Dong et~al.(2021)Dong, Cordonnier, and Loukas]{dong2021pureattention}
Yihe Dong, Jean-Baptiste Cordonnier, and Andreas Loukas.
\newblock Attention is not all you need: Pure attention loses rank doubly
  exponentially with depth.
\newblock In \emph{Int. Conf. Mach. Learn. (ICML)}, pages 2793--2803, 2021.

\bibitem[Fan et~al.(2025)Fan, Fan, Hu, Ding, Di, Yi, Pagnucco, and
  Song]{fan2025manta}
Lei Fan, Dongdong Fan, Zhiguang Hu, Yiwen Ding, Donglin Di, Kai Yi, Maurice
  Pagnucco, and Yang Song.
\newblock {MANTA}: A large-scale multi-view and visual-text anomaly detection
  dataset for tiny objects.
\newblock In \emph{CVPR}, pages 25518--25527, 2025.

\bibitem[Gudovskiy et~al.(2022)Gudovskiy, Ishizaka, and
  Kozuka]{gudovskiy2022cflow}
Denis Gudovskiy, Shun Ishizaka, and Kazuki Kozuka.
\newblock {CFLOW-AD}: Real-time unsupervised anomaly detection with
  localization via conditional normalizing flows.
\newblock In \emph{Proceedings of the IEEE/CVF Winter Conference on
  Applications of Computer Vision}, pages 98--107, 2022.

\bibitem[Guo et~al.(2023)Guo, Lu, Jia, Zhang, and Li]{guo2023recontrast}
Jia Guo, Shuai Lu, Lize Jia, Weihang Zhang, and Huiqi Li.
\newblock {ReContrast}: Domain-specific anomaly detection via contrastive
  reconstruction.
\newblock In \emph{NeurIPS}, pages 10721--10740, 2023.

\bibitem[Guo et~al.(2025{\natexlab{a}})Guo, Lu, Fan, Li, Di, Song, Zhang, Zhu,
  Yan, Chen, Li, and Liao]{guo2025dinomaly2}
Jia Guo, Shuai Lu, Lei Fan, Zelin Li, Donglin Di, Yang Song, Weihang Zhang,
  Wenbing Zhu, Hong Yan, Fang Chen, Huiqi Li, and Hongen Liao.
\newblock Dinomaly2: A unified framework for unsupervised image anomaly
  detection.
\newblock arXiv preprint arXiv:2510.17611, 2025{\natexlab{a}}.

\bibitem[Guo et~al.(2025{\natexlab{b}})Guo, Lu, Zhang, Chen, Li, and
  Liao]{guo2025dinomaly}
Jia Guo, Shuai Lu, Weihang Zhang, Fang Chen, Huiqi Li, and Hongen Liao.
\newblock Dinomaly: The less is more philosophy in multi-class unsupervised
  anomaly detection.
\newblock In \emph{CVPR}, pages 20405--20415, 2025{\natexlab{b}}.

\bibitem[Haber and Ruthotto(2018)]{haber2017stable}
Eldad Haber and Lars Ruthotto.
\newblock Stable architectures for deep neural networks.
\newblock \emph{Inverse Problems}, 34\penalty0 (1):\penalty0 014004, 2018.

\bibitem[He et~al.(2024)He, Bai, Zhang, He, Chen, Gan, Wang, Li, Tian, and
  Xie]{he2024mambaad}
Haoyang He, Yuhu Bai, Jiangning Zhang, Qingdong He, Hongxu Chen, Zhenye Gan,
  Chengjie Wang, Xiangtai Li, Guanzhong Tian, and Lei Xie.
\newblock {MambaAD}: Exploring state space models for multi-class unsupervised
  anomaly detection.
\newblock In \emph{NeurIPS}, pages 71162--71187, 2024.

\bibitem[Jacobs et~al.(2025)Jacobs, Fel, Hakim, Brondetta, Ba, and
  Keller]{jacobs2025blockrecurrent}
Mozes Jacobs, Thomas Fel, Richard Hakim, Alessandra Brondetta, Demba Ba, and
  T.~Andy Keller.
\newblock Block-recurrent dynamics in vision transformers.
\newblock arXiv preprint arXiv:2512.19941, 2025.

\bibitem[Jezek et~al.(2021)Jezek, Jonak, Burget, Dvorak, and
  Skotak]{jezek2021mpdd}
Stepan Jezek, Martin Jonak, Radim Burget, Pavel Dvorak, and Milos Skotak.
\newblock Deep learning-based defect detection of metal parts: Evaluating
  current methods in complex conditions.
\newblock In \emph{International Congress on Ultra Modern Telecommunications
  and Control Systems and Workshops (ICUMT)}, pages 66--71, 2021.

\bibitem[Jiang et~al.(2022)Jiang, Liu, Wang, Nie, Wu, Liu, Wang, and
  Zheng]{jiang2022softpatch}
Xi Jiang, Jianlin Liu, Jinbao Wang, Qiang Nie, Kai Wu, Yong Liu, Chengjie Wang,
  and Feng Zheng.
\newblock {SoftPatch}: Unsupervised anomaly detection with noisy data.
\newblock In \emph{NeurIPS}, pages 15433--15445, 2022.

\bibitem[Kumar(2025)]{krishnakumar2025cliffplateau}
Anantha Padmanaban~Krishna Kumar.
\newblock Mechanisms of non-monotonic scaling in vision transformers.
\newblock arXiv preprint arXiv:2511.21635, 2025.

\bibitem[Li et~al.(2021)Li, Sohn, Yoon, and Pfister]{li2021cutpaste}
Chun-Liang Li, Kihyuk Sohn, Jinsung Yoon, and Tomas Pfister.
\newblock {CutPaste}: Self-supervised learning for anomaly detection and
  localization.
\newblock In \emph{CVPR}, pages 9664--9674, 2021.

\bibitem[Liu et~al.(2023)Liu, Zhou, Xu, and Wang]{liu2023simplenet}
Zhikang Liu, Yiming Zhou, Yuansheng Xu, and Zilei Wang.
\newblock {SimpleNet}: A simple network for image anomaly detection and
  localization.
\newblock In \emph{CVPR}, pages 20402--20411, 2023.

\bibitem[Loshchilov and Hutter(2019)]{loshchilov2019adamw}
Ilya Loshchilov and Frank Hutter.
\newblock Decoupled weight decay regularization.
\newblock In \emph{ICLR}, 2019.

\bibitem[Mishra et~al.(2021)Mishra, Verk, Fornasier, Piciarelli, and
  Foresti]{mishra2021btad}
Pankaj Mishra, Riccardo Verk, Daniele Fornasier, Claudio Piciarelli, and
  Gian~Luca Foresti.
\newblock {VT-ADL}: A vision transformer network for image anomaly detection
  and localization.
\newblock In \emph{IEEE International Symposium on Industrial Electronics
  (ISIE)}, pages 1--6, 2021.

\bibitem[Noci et~al.(2022)Noci, Anagnostidis, Biggio, Orvieto, Singh, and
  Lucchi]{noci2022signalprop}
Lorenzo Noci, Sotiris Anagnostidis, Luca Biggio, Antonio Orvieto, Sidak~Pal
  Singh, and Aur{\'e}lien Lucchi.
\newblock Signal propagation in transformers: Theoretical perspectives and the
  role of rank collapse.
\newblock In \emph{NeurIPS}, pages 27198--27211, 2022.

\bibitem[Oquab et~al.(2024)Oquab, Darcet, Moutakanni, Vo, Szafraniec, Khalidov,
  Fernandez, Haziza, Massa, El-Nouby, Assran, Ballas, Galuba, Howes, Huang, Li,
  Misra, Rabbat, Sharma, Synnaeve, Xu, J{\'e}gou, Mairal, Labatut, Joulin, and
  Bojanowski]{oquab2024dinov2}
Maxime Oquab, Timoth{\'e}e Darcet, Th{\'e}o Moutakanni, Huy Vo, Marc
  Szafraniec, Vasil Khalidov, Pierre Fernandez, Daniel Haziza, Francisco Massa,
  Alaaeldin El-Nouby, Mahmoud Assran, Nicolas Ballas, Wojciech Galuba, Russell
  Howes, Po-Yao Huang, Shang-Wen Li, Ishan Misra, Michael Rabbat, Vasu Sharma,
  Gabriel Synnaeve, Hu Xu, Herv{\'e} J{\'e}gou, Julien Mairal, Patrick Labatut,
  Armand Joulin, and Piotr Bojanowski.
\newblock {DINOv2}: Learning robust visual features without supervision.
\newblock \emph{Transactions on Machine Learning Research}, 2024.

\bibitem[Papyan et~al.(2020)Papyan, Han, and Donoho]{papyan2020neuralcollapse}
Vardan Papyan, X.~Y. Han, and David~L. Donoho.
\newblock Prevalence of neural collapse during the terminal phase of deep
  learning training.
\newblock \emph{Proceedings of the National Academy of Sciences}, 117\penalty0
  (40):\penalty0 24652--24663, 2020.

\bibitem[Patyn and Deconinck(2024)]{patyn2024koopman}
Christophe Patyn and Geert Deconinck.
\newblock A concise introduction to {Koopman} operator theory and the {Extended
  Dynamic Mode Decomposition}.
\newblock arXiv preprint arXiv:2410.02766, 2024.

\bibitem[Ranzinger et~al.(2024)Ranzinger, Heinrich, Kautz, and
  Molchanov]{ranzinger2024amradio}
Mike Ranzinger, Greg Heinrich, Jan Kautz, and Pavlo Molchanov.
\newblock {AM-RADIO}: Agglomerative vision foundation model -- reduce all
  domains into one.
\newblock In \emph{Proceedings of the IEEE/CVF Conference on Computer Vision
  and Pattern Recognition}, pages 12490--12500, 2024.

\bibitem[Ravi et~al.(2025)Ravi, Gabeur, Hu, Hu, Ryali, Ma, Khedr, R{\"a}dle,
  Rolland, Gustafson, Mintun, Pan, Alwala, Carion, Wu, Girshick, Doll{\'a}r,
  and Feichtenhofer]{ravi2025sam2}
Nikhila Ravi, Valentin Gabeur, Yuan-Ting Hu, Ronghang Hu, Chaitanya Ryali,
  Tengyu Ma, Haitham Khedr, Roman R{\"a}dle, Chloe Rolland, Laura Gustafson,
  Eric Mintun, Junting Pan, Kalyan~Vasudev Alwala, Nicolas Carion, Chao-Yuan
  Wu, Ross Girshick, Piotr Doll{\'a}r, and Christoph Feichtenhofer.
\newblock {SAM} 2: Segment anything in images and videos.
\newblock In \emph{ICLR}, 2025.

\bibitem[Roth et~al.(2022)Roth, Pemula, Zepeda, Sch{\"o}lkopf, Brox, and
  Gehler]{roth2022patchcore}
Karsten Roth, Latha Pemula, Joaquin Zepeda, Bernhard Sch{\"o}lkopf, Thomas
  Brox, and Peter Gehler.
\newblock Towards total recall in industrial anomaly detection.
\newblock In \emph{CVPR}, pages 14318--14328, 2022.

\bibitem[Shopkhoev et~al.(2025)Shopkhoev, Ali, Zhussip, Malykh, Lefkimmiatis,
  Komodakis, and Zagoruyko]{shopkhoev2025replaceme}
Dmitriy Shopkhoev, Ammar Ali, Magauiya Zhussip, Valentin Malykh, Stamatios
  Lefkimmiatis, Nikos Komodakis, and Sergey Zagoruyko.
\newblock Replaceme: Network simplification via depth pruning and transformer
  block linearization.
\newblock In \emph{NeurIPS}, pages 22092--22122, 2025.

\bibitem[Wang et~al.(2024)Wang, Zhu, Gao, Gan, Zhang, Gu, Qian, Chen, and
  Ma]{wang2024realiad}
Chengjie Wang, Wenbing Zhu, Bin-Bin Gao, Zhenye Gan, Jiangning Zhang, Zhihao
  Gu, Shuguang Qian, Mingang Chen, and Lizhuang Ma.
\newblock {Real-IAD}: A real-world multi-view dataset for benchmarking
  versatile industrial anomaly detection.
\newblock In \emph{CVPR}, pages 22883--22892, 2024.

\bibitem[Wieler and Hahn(2007)]{wieler2007dagm}
M. Wieler and T. Hahn.
\newblock Weakly supervised learning for industrial optical inspection.
\newblock DAGM Symposium, Competition Track, 2007.

\bibitem[You et~al.(2022)You, Cui, Shen, Yang, Lu, Zheng, and Le]{you2022uniad}
Zhiyuan You, Lei Cui, Yujun Shen, Kai Yang, Xin Lu, Yu Zheng, and Xinyi Le.
\newblock A unified model for multi-class anomaly detection.
\newblock In \emph{NeurIPS}, pages 4571--4584, 2022.

\bibitem[Zavrtanik et~al.(2021)Zavrtanik, Kristan, and Sko{\v
  c}aj]{zavrtanik2021draem}
Vitjan Zavrtanik, Matej Kristan, and Danijel Sko{\v c}aj.
\newblock {DRAEM} -- a discriminatively trained reconstruction embedding for
  surface anomaly detection.
\newblock In \emph{ICCV}, pages 8330--8339, 2021.

\bibitem[Zhang et~al.(2025)Zhang, Chen, Wang, Wang, Liu, Li, Yang, and
  Tao]{zhang2025vitad}
Jiangning Zhang, Xuhai Chen, Yabiao Wang, Chengjie Wang, Yong Liu, Xiangtai Li,
  Ming-Hsuan Yang, and Dacheng Tao.
\newblock Exploring plain {ViT} features for multi-class unsupervised visual
  anomaly detection.
\newblock \emph{Computer Vision and Image Understanding}, 253:\penalty0 104308,
  2025.

\bibitem[Zhang et~al.(2023)Zhang, Li, Li, Huang, Shan, and
  Chen]{zhang2023destseg}
Xuan Zhang, Shiyu Li, Xi Li, Ping Huang, Jiulong Shan, and Ting Chen.
\newblock {DeSTSeg}: Segmentation guided denoising student-teacher for anomaly
  detection.
\newblock In \emph{CVPR}, pages 3914--3923, 2023.

\bibitem[Zou et~al.(2022)Zou, Jeong, Pemula, Zhang, and Dabeer]{zou2022visa}
Yang Zou, Jongheon Jeong, Latha Pemula, Dongqing Zhang, and Onkar Dabeer.
\newblock {SPot-the-Difference} self-supervised pre-training for anomaly
  detection and segmentation.
\newblock In \emph{ECCV}, pages 392--408, 2022.

\end{thebibliography}
}

\end{document}